%% file: ijcai26.tex
\documentclass{article}
\newcommand{\tuner}{\texttt{CDS4RAG}}
\usepackage{ijcai26}

\usepackage{times}
\usepackage{soul}
\usepackage{url}
\usepackage[hidelinks]{hyperref}
\usepackage[utf8]{inputenc}
\usepackage[small]{caption}
\usepackage{adjustbox}
\usepackage{graphicx}
\usepackage{amsthm}
\usepackage{booktabs}
\usepackage[switch]{lineno}
\usepackage{latexsym}
\usepackage{amsmath,amsfonts}
\usepackage{pgfplots}
\usepackage{colortbl}
\usepackage{tcolorbox}
\usepackage{standalone}
\usepackage{subfig}
\usepackage{tikz}  
\usetikzlibrary{backgrounds}

\usepackage{microtype}

\usepackage{inconsolata}

\usepackage{algorithm}
\usepackage{algpseudocode}
\usepackage{multirow}
\usepackage{enumitem}

\definecolor{rwth-blue}{RGB}{0,84,159}
\definecolor{rwth-red}{RGB}{204,7,30}
\definecolor{rwth-grey}{RGB}{87,87,87}

\newcommand{\repolink}[2]{#1~\ref{#2}}

\title{CDS4RAG: Cyclic Dual-Sequential Hyperparameter Optimization for RAG}

\author{
Pengzhou Chen$^1$\footnote{Pengzhou Chen is also supervised in the IDEAS Lab.}
\and
Tao Chen$^2$\footnote{Tao Chen is the corresponding author.}\\
\affiliations
$^1$School of Computer Science and Engineering, UESTC, Chengdu, China\\
$^2$IDEAS Lab, University of Birmingham, Birmingham, UK\\
\emails
t.chen@bham.ac.uk
}

\begin{document}

\maketitle

\begin{abstract}
Retrieval-Augmented Generation (RAG) is sensitive to the vast hyperparameters of the retriever and generator, yet optimizing them using given queries is a challenging task due to the complex interactions and expensive evaluation costs. Existing algorithms are ineffective and slow in convergence, since they often treat RAG as a monolithic black box or only optimize partial hyperparameters. In this paper, we propose \tuner, a framework that optimizes the full RAG hyperparameters using given queries via a new cyclic dual-sequential formulation. \tuner~is special in the sense that it distinguishes the hyperparameters of the retriever and generator, cyclically optimizing them in turn. Such a paradigm allows us to design fine-grained within-cycle budget provision and expedite the optimization via cross-cycle seeding when optimizing the generator. \tuner~is also an algorithm-agnostic framework that can be paired with diverse general algorithms. Through experiments on four common benchmarks and two backbone LLMs, we reveal that \tuner~considerably boosts the vanilla algorithms in 21/24 cases while significantly outperforming state-of-the-art algorithms in all cases with up to $1.54\times$ improvements of generation quality and better speedup.

\end{abstract}

\input{latex/introduction}

\input{latex/related_work}

\input{latex/preliminary}

\input{latex/methodology}

\input{latex/experiments}

\input{latex/results}

\input{latex/conclusion}

\section*{Acknowledgment}
This work was supported by a NSFC Grant (62372084).

\bibliographystyle{named}
\bibliography{ijcai26}

\input{appendix-content}

\end{document}

%% file: latex/introduction.tex
\section{Introduction}

\begin{figure}[t!]
    \centering
    \includegraphics[width=\linewidth]{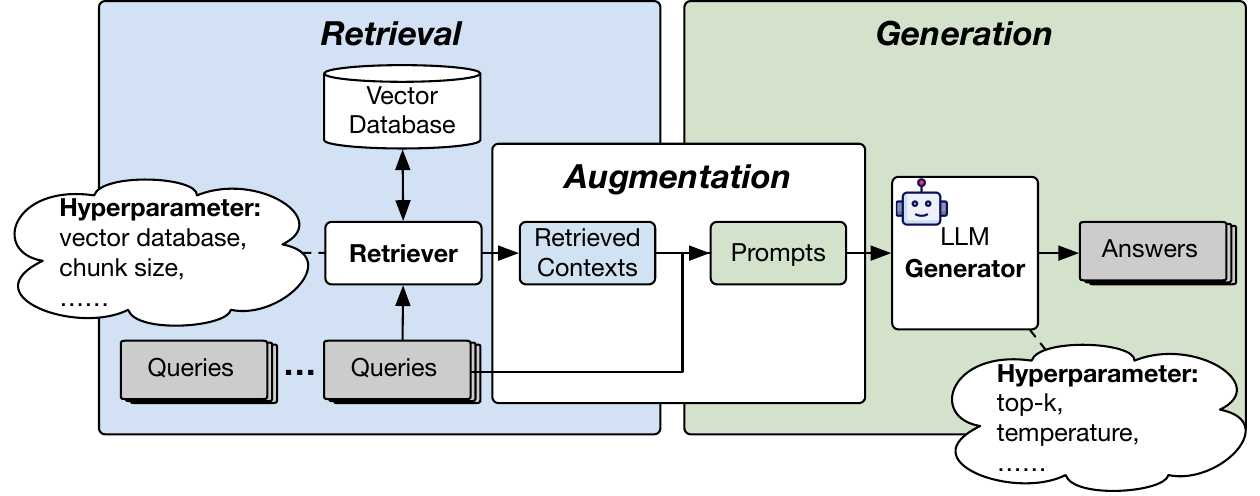}
    \caption{General workflow and hyperparameters of RAG.}
    \label{fig:rag-example}
\end{figure}

Retrieval-Augmented Generation (RAG) has emerged as a dominant paradigm for enhancing Large Language Models (LLMs) based on external knowledge~\cite{MsRAG,lightrag}, serving as a foundation in many AI applications~\cite{AI4Contracts,DBLP:conf/acl/NiuWZXSZS024}. As from Figure~\ref{fig:rag-example}, typically the RAG contains a retriever (i.e., the retrieval stage linked with a vector/graph database) and a generator (i.e., the generation stage by a LLM): for a given batch of queries, the retriever retrieves relevant contexts that consolidate the query prompts for the generator to produce better answers.

Like many other systems~\cite{DBLP:journals/tosem/ChenLBY18,DBLP:journals/tse/ChenGC25,10.1145/3803859}, the effectiveness of RAG in handling queries highly depends on its hyperparameter values~\cite{autorag-hp}, i.e., those for the retrieval (e.g., chunk size) and the generation stage (e.g., generation temperature), thus the their optimization is crucial. However, optimizing RAG hyperparameters using given queries is challenging, because:  

\begin{itemize}
    \item With the combined retriever and generator in RAG, the dimension of tunable hyperparameters increases dramatically with more complex interactions, making optimization difficult~\cite{Faster,Rag_and_Roll}.
    \item Evaluating the quality of a hyperparameter configurations as the feedback under full retrieval-generation round can be expensive, e.g., configuration evaluation on a RAG with \texttt{Llama-3.1-8B} over the \textit{Agriculture} benchmark can take up to $30$ minutes~\cite{lightrag}. 
\end{itemize}

To that end, many algorithms have been proposed. For example, there exist general algorithms for tuning hyperparameters~\cite{TPE,Cowen-Rivers2022-HEBO}; specific ones such as \texttt{AutoRAG-HP} \cite{autorag-hp} and \texttt{METIS} \cite{Metis} for RAG that focus on the entire retrieval-generation process as a whole; as well as others that solely perform hyperparameter optimization/HPO related to the prompt or the retriever of RAG~\cite{GEPA,Adaptiverag}. Yet, those algorithms are limited in the sense that they merely treat RAG as a monolithic black box or restrict themselves only to the hyperparameters at a certain stage. This ignores the highly structural internal information of RAG, causing slow convergence and devastating results.

In this paper, we take a different perspective to bridge these gaps: drawing on our understandings and empirical experiments, we observe that the retriever and generator in RAG are interrelated but also partially separable, each plays different roles that determine the final generation outcomes of RAG. As such, we hypothesize that \textit{``if the hyperparameters of retriever and generator can be optimized loosely, in a way that they maintain merely the basic interactions, then it should yield better results using less budget.''} Thus, we propose \tuner, a cyclic dual-sequential framework that heuristically optimizes RAG's hyperparameters using given queries. What make \tuner~special is that it distinguishes the retriever and generator in RAG, optimizing their corresponding hyperparameters sequentially with different objectives, but doing so in a cyclic manner\footnote{A cycle means the completion of one sequence of optimizing retriever and then generator.}, permitting a fine-grained budget provision and seeding\footnote{Seeding means initialize an optimization with pre-defined hyperparameter configurations to start working with, enabling a warm-start.} therein. This allows \tuner~to \textbf{focus on the sliced hyperparameter space}, hence mitigating the issue of complex hyperparameter interactions, while \textbf{relieving the optimization from evaluating the costly full retrieval-generation} via a good proportion of retrieval-only evaluations.

Since the cyclic dual-sequential hyperparameter optimization is compatible with any general algorithms, \tuner~is also algorithm-agnostic. Specifically, our contributions are:

\begin{itemize}
    \item We newly reformulate the RAG hyperparameter optimization as a cyclic dual-sequential problem.
    \item We present a way to sequentially optimize the hyperparameters of the retriever and generator for retrieval and generation quality, respectively, in a {cyclic manner}.
    \item To achieve fine-grained control over the budget provisioning of hyperparameter optimization between retrieval and generation within a cycle, we propose an adaptive mechanism that stops optimizing the retriever when it detects a retrieval plateau, while capping that for the generator with a fixed amount of evaluations. This reduces the costly full retrieval-generation evaluations needed. 
    \item To expedite the optimization with a warm-start, we design a cross-cycle seeding, such that promising generation hyperparameter configurations in previous cycles, based on knowledge retrieved under different retrieval configurations, are reused in the current cycle for the generator.
\end{itemize}

Experimental results on four common benchmarks and two backbone LLMs reveal that \tuner~can considerably boost general algorithms in 21/24 cases while significantly outperforms other state-of-the-art hyperparameter optimization algorithms for RAG in all cases with quality improvement of up to $1.54\times$ and better speedup. All data, code, and appendices can be accessed at: \texttt{\textcolor{blue}{\href{https://github.com/ideas-labo/cds4rag}{https://github.com/ideas-labo/cds4rag}}}.





%% file: latex/related_work.tex
\section{Related Work}


\subsection{General Hyperparameter Optimization}


Over the decades, general algorithms like Bayesian Optimization (\texttt{BO})~\cite{BO_NIPS} and its variants~\cite{BO_NIPS,TPE,openbox,DBLP:conf/icse/ChenChen26,DBLP:conf/icse/LiX0WT20} have been proposed for hyperparameter optimization. Those methods rely on a surrogate model~\cite{DBLP:journals/tse/GongCB25,DBLP:conf/icse/XiangChen26,DBLP:conf/sigsoft/Gong023,DBLP:journals/pacmse/Gong024} to predict the next evaluated configuration, balancing the exploration and exploitation with reduced evaluation cost. State-of-the-art algorithms like \texttt{HEBO} \cite{Cowen-Rivers2022-HEBO} have also been developed to handle more complex and noisy search spaces robustly. Other metaheuristics without a model also exist~\cite{DBLP:conf/kbse/XiongC25,chen2024mmo,DBLP:conf/sigsoft/0001L21,DBLP:conf/sigsoft/0001L24,DBLP:conf/icse/Ye0L25,DBLP:conf/wcre/Chen22}. 


However, the above often do not use any specific knowledge/information of the algorithm/system to tune, which can waste the valuable information of the known structure in RAG.

\subsection{Full Hyperparameter Optimization for RAG}

There are algorithms designed/tailored for optimizing RAG hyperparameters using given queries. Among others, \texttt{RayTune}~\cite{RayTune} tailors the general black-box optimization algorithms TPE to optimize the RAG hyperparameters. \texttt{AutoRAG-HP} \cite{autorag-hp} utilizes Hierarchical Multi-Armed Bandits to optimize the RAG hyperparameters. While effective for discrete/combinatorial hyperparameters, it struggles with continuous ones. \texttt{METIS} \cite{Metis} is a multi-objective optimization approach to balance quality and latency in RAG, but it can struggle under high-dimensionality. 



Yet, again those works often treat the RAG pipeline as a monolithic black box, relying on the evaluation of the full retrieval-generation rounds throughout the optimization.


\subsection{Partial Hyperparameter Optimization for RAG}

A parallel thread of research focuses on optimizing the hyperparameters of a specific stage for RAG, e.g., \texttt{GEPA} \cite{GEPA} and \texttt{MIPROv2} \cite{Miprov2} use LLMs to iteratively refine instructions for the prompt at the generator, but overlook the hyperparameters of the retriever (e.g., chunk size) that dictate the underlying context. In contrast, methods like \texttt{AdaptiveRAG} \cite{Adaptiverag} dynamically adjust retrieval intensity only (e.g., top-$k$) based on query complexity. Other studies justify the importance of hyperparameters at diverse stages in RAG~\cite{Analysis}. 

However, focusing on a single stage in RAG hyperparameter optimization can miss the valuable configurations that can only be found when different stages are considered together. Yet, the complex interaction and expensive evaluation make a direct joint optimization complex. This work tackles exactly such via cyclic dual-sequential hyperparameter optimization.



%% file: latex/preliminary.tex
\section{Problem Formulation and Motivation}



\subsection{Classic RAG Hyperparameter Optimization}

A hyperparameter in RAG can be a continuous, enumerated, or categorical variable. Traditionally, hyperparameter optimization for RAG works in a combined hyperparameter space of configurations $\Omega = {\Phi} \times {\Theta}$ from that of the retriever ${\Phi}$ and generator ${\Theta}$. Formally, given a query batch $q\in\mathcal{Q}$ and a knowledge corpus $\mathcal{D}$, this can be expressed as:
\begin{equation}
   \arg\max_{\boldsymbol{\omega} \in \Omega}  \left[ \mathbf{M}(f(\mathcal{Q}, \mathcal{D},\boldsymbol{\omega}), \mathcal{G}^*) \right]
    \label{eq:original}
\end{equation}
whereby $f$ is the RAG system; $\mathbf{M}$ is the end-to-end performance metric over $\mathcal{Q}$; $\mathcal{G}^*$ is the ground truth for answering all the queries. The goal is to jointly find the best hyperparameter configuration for both retriever and generator using $\mathcal{Q}$, hence $\mathbf{M}$, e.g., the F1-score of the generated answers, is maximized.


Albeit straightforward and widely adopted in existing work, this formulation forces one to work on the full combined space of retriever and generator under the costly full retrieval-generation evaluations, which can be slow and ineffective \cite{Analysis,autorag-hp,Faster}.



\begin{figure}[t!]
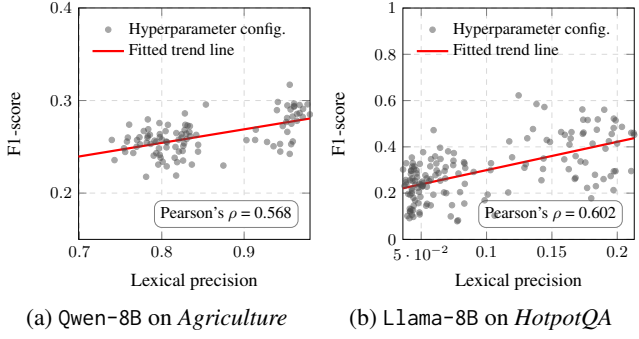

\centering
\subfloat[\texttt{Qwen-8B} on \textit{Agriculture}]{
\includestandalone[width=0.48\linewidth]{picture/agriculture_correlation}
}
\subfloat[\texttt{Llama-8B} on \textit{HotpotQA}]{
\includestandalone[width=0.48\linewidth]{picture/hotpot_correlation}
}
 \caption{Exampled correlations between the quality of retrieval and generation over different RAG hyperparameter configurations under some queries. More cases can be found at \repolink{Appendix}{app:cases}.}
    \label{fig:agri_correlation}
\end{figure}

\subsection{Cyclic Dual-Sequential Hyperparameter Optimization for RAG}

While we can intuitively understand that the retriever and generator are two independent components in RAG, there is also stronger empirical evidence to motivate this work. From Figure~\ref{fig:agri_correlation}, we see that the quality of contexts retrieved and answers generated under some queries, i.e., lexical precision and F1-score, commonly exhibit good positively monotonic correlation, which is also evidenced by prior studies \cite{correlation,Analysis}. Further, we found that the evaluation cost of retrievers is often $\approx50\%$ cheaper than that of the generator. This motivates us to reformulate Equation~\ref{eq:original} as:
\begin{subequations}
\label{eq:two-stage-optimization}
\begin{flalign}
&   \arg\max_{(\boldsymbol{\phi},\boldsymbol{\theta}) \in \{(\boldsymbol{\phi}^*_1,\boldsymbol{\theta}^*_1),\dots,(\boldsymbol{\phi}^*_T,\boldsymbol{\theta}^*_T)\}}  \left[ \mathbf{M}(f(\mathcal{Q}, \mathcal{D},\boldsymbol{\phi},\boldsymbol{\theta}), \mathcal{G}^*) \right] \label{eq:stage1} \\
& \text{\>s.t. } \boldsymbol{\theta}^*_t \in \arg\max_{\boldsymbol{\theta} \in {\Theta}}  \left[ \mathbf{M}\left( f_g(\mathcal{Q}, \mathcal{C}^*_t, \boldsymbol{\theta}), \mathcal{G}^* \right) \right] \label{eq:stage2} \\
& \text{\;\>\quad} \boldsymbol{\phi}^*_t \in \arg\max_{\boldsymbol{\phi} \in {\Phi}}  \left[ \mathbf{P}(f_r(\mathcal{Q}, \mathcal{D}, \boldsymbol{\phi}), \mathcal{R}^*) \right] \label{eq:stage3}\\
& \text{\;\>\quad} \mathcal{C}^*_t=f_r(\mathcal{Q}, \mathcal{D}, \boldsymbol{\phi}^*_t)\label{eq:stage4}
\end{flalign}
\end{subequations}
where $\boldsymbol{\omega}=(\boldsymbol{\phi},\boldsymbol{\theta})$; $f_g$ and $f_r$ are the generator and retriever, respectively; $\boldsymbol{\phi}^*_t$ and $\boldsymbol{\theta}^*_t$ are their hyperparameter configurations optimized at cycle $t$ out of $T$ cycles. $\mathbf{P}$ is the quality metric for the retriever, e.g., the lexical precision; $\mathcal{R}^*$ is the ground truth of the contexts for all queries. Using $\mathcal{Q}$, we seek to tackle Equation~\ref{eq:stage1} by exploring the best hyperparameter configuration $\boldsymbol{\phi}_t^*$ for retriever under $\mathbf{P}$ first, and then use the retrieved contexts $\mathcal{C}_t^*$ under $\boldsymbol{\phi}_t^*$ to jointly optimize and find the optimal generator configuration, leading to the best $\boldsymbol{\theta}^*_t$ on $\mathbf{M}$ thereunder at cycle $t$. This process is then repeated in a cyclic manner, eventually resulting in the best global $(\boldsymbol{\phi}^*,\boldsymbol{\theta}^*)$ for $\mathcal{Q}$. \textbf{Note that the above is not a classic bi-level/block-coordinate optimization, but rather a newly emerged, cyclic dual-sequential optimization at the same level for RAG.} This is because in those classic paradigms, each step often optimizes one block/level of variables with respect to the same global objective~\cite{DBLP:journals/tec/SinhaMD18}. In our case, the cyclic dual-sequential formula uses distinct stage-specific objectives (Equations~\ref{eq:stage2} and~\ref{eq:stage3}), which are implicitly related to each other in the ultimate goal (Equation~\ref{eq:stage1}).

%% file: picture/agriculture_correlation.tex
\begin{tikzpicture}
\begin{axis}[
    width=6cm,
    height=6cm,
    xlabel={{Lexical precision}},
    ylabel={{F1-score}},
    xmin=0.699, xmax=0.980,
    ymin=0.150, ymax=0.4,
    grid=major,
    grid style={dashed, gray!30},
    legend pos=north west,
    legend cell align={left},
    legend style={font=\small,draw=none,fill=none},
    tick label style={font=\small},
    label style={font=\normalsize},
]

\addplot[
    only marks,
    mark=*,
    mark size=1.5pt,
    color=black!70,
    opacity=0.5,
] coordinates {
(0.785574,0.263055)
(0.927837,0.255248)
(0.943861,0.244868)
(0.828524,0.273807)
(0.827139,0.253804)
(0.796509,0.248273)
(0.698825,0.244533)
(0.826157,0.250936)
(0.960644,0.296777)
(0.971563,0.290931)
(0.955350,0.242361)
(0.824176,0.263613)
(0.783095,0.261889)
(0.793417,0.243196)
(0.805972,0.228896)
(0.817682,0.264851)
(0.825876,0.235127)
(0.830160,0.258467)
(0.979890,0.285002)
(0.739316,0.257429)
(0.797541,0.241330)
(0.820813,0.244916)
(0.760408,0.273973)
(0.765415,0.243344)
(0.824043,0.259928)
(0.790183,0.240450)
(0.960436,0.293318)
(0.955365,0.317060)
(0.951159,0.276805)
(0.853557,0.295675)
(0.834155,0.272313)
(0.958852,0.282631)
(0.780983,0.217636)
(0.769963,0.257335)
(0.807467,0.273760)
(0.939810,0.289582)
(0.801079,0.233900)
(0.946853,0.272434)
(0.797144,0.275689)
(0.956376,0.290351)
(0.782621,0.259759)
(0.844190,0.252210)
(0.913347,0.257413)
(0.976981,0.295906)
(0.902816,0.257188)
(0.817439,0.218947)
(0.827726,0.268176)
(0.964182,0.285488)
(0.967318,0.286456)
(0.836517,0.230826)
(0.943651,0.253095)
(0.769410,0.250046)
(0.778125,0.267056)
(0.831364,0.272993)
(0.806348,0.254878)
(0.800263,0.257859)
(0.742611,0.235635)
(0.935705,0.250598)
(0.825342,0.245510)
(0.793335,0.255552)
(0.815646,0.238150)
(0.797581,0.263714)
(0.805718,0.249369)
(0.795749,0.230815)
(0.959100,0.255789)
(0.783615,0.268987)
(0.798534,0.238211)
(0.914129,0.268514)
(0.960953,0.294098)
(0.805552,0.255669)
(0.952455,0.266132)
(0.834531,0.261414)
(0.754155,0.248452)
(0.757709,0.255656)
(0.772928,0.260331)
(0.844016,0.244813)
(0.816616,0.249883)
(0.755550,0.244450)
(0.829774,0.264232)
(0.940920,0.298433)
(0.952141,0.281107)
(0.967318,0.295146)
(0.957153,0.252814)
(0.783052,0.279595)
(0.781733,0.254531)
(0.797663,0.260444)
(0.758787,0.259991)
(0.773665,0.252665)
(0.833530,0.265936)
(0.786468,0.253137)
(0.828052,0.286071)
(0.804609,0.266938)
(0.820542,0.257800)
(0.952505,0.294687)
(0.971113,0.282328)
(0.818640,0.233412)
(0.964182,0.278090)
(0.957440,0.274935)
(0.791178,0.256645)
(0.874806,0.229757)
(0.841917,0.262900)
};
\addlegendentry{Hyperparameter config.}

\addplot[
    color=red,
    line width=1.2pt,
] coordinates {
(0.698825,0.239446) (0.979890,0.280600)
};
\addlegendentry{Fitted trend line}

\node[
    draw=gray,
    fill=white,
    rounded corners,
    font=\small,
    anchor=south east
] at (rel axis cs:0.95,0.05) {Pearson's $\rho$ = 0.568};

\end{axis}
\end{tikzpicture}

%% file: picture/hotpot_correlation.tex
\begin{tikzpicture}
\begin{axis}[
    width=6cm,
    height=6cm,
    xlabel={{Lexical precision}},
    ylabel={{F1-score}},
    xmin=0.036, xmax=0.213,
    ymin=0, ymax=1.00,
   grid=major,
    grid style={dashed, gray!30},
    legend pos=north west,
    legend cell align={left},
    legend style={font=\small,draw=none,fill=none},
    tick label style={font=\small},
    label style={font=\normalsize},
]

\addplot[
    only marks,
    mark=*,
    mark size=1.5pt,
    color=black!70,
    opacity=0.5,
] coordinates {
(0.038678,0.236874)
(0.191003,0.392256)
(0.040084,0.196480)
(0.044287,0.145868)
(0.046818,0.268798)
(0.044780,0.315389)
(0.173589,0.456385)
(0.198643,0.294860)
(0.098516,0.169020)
(0.048867,0.107989)
(0.162867,0.303448)
(0.054516,0.154000)
(0.041573,0.119752)
(0.039403,0.101298)
(0.042852,0.353944)
(0.073569,0.147893)
(0.060911,0.277947)
(0.078510,0.284774)
(0.050921,0.231852)
(0.082719,0.333597)
(0.050288,0.226344)
(0.183642,0.230034)
(0.173956,0.517254)
(0.038220,0.256834)
(0.063302,0.286333)
(0.185071,0.518520)
(0.045453,0.236739)
(0.056174,0.252175)
(0.041407,0.255952)
(0.078186,0.082443)
(0.046421,0.298838)
(0.059312,0.193261)
(0.049627,0.236233)
(0.042864,0.106436)
(0.043555,0.329727)
(0.048667,0.240071)
(0.043213,0.315040)
(0.042794,0.199105)
(0.045379,0.168275)
(0.061791,0.298401)
(0.072615,0.099473)
(0.043259,0.253111)
(0.067421,0.202982)
(0.090125,0.291648)
(0.041549,0.192762)
(0.074685,0.340286)
(0.144255,0.572204)
(0.056939,0.315768)
(0.157067,0.464036)
(0.171902,0.336842)
(0.074511,0.320940)
(0.180090,0.177689)
(0.061608,0.328206)
(0.163437,0.409829)
(0.210683,0.295357)
(0.054258,0.263698)
(0.103421,0.329164)
(0.069178,0.200035)
(0.053783,0.145596)
(0.057526,0.346810)
(0.046659,0.285040)
(0.083356,0.240166)
(0.124533,0.622462)
(0.168145,0.359990)
(0.181109,0.413919)
(0.193573,0.424091)
(0.077564,0.078573)
(0.068746,0.312373)
(0.070590,0.380609)
(0.049414,0.184007)
(0.064500,0.396182)
(0.044061,0.179997)
(0.212519,0.460324)
(0.119555,0.354802)
(0.142000,0.255598)
(0.194959,0.273594)
(0.048657,0.116529)
(0.164986,0.463617)
(0.117376,0.423659)
(0.042286,0.297601)
(0.195753,0.455760)
(0.043244,0.263941)
(0.057322,0.147190)
(0.054343,0.227700)
(0.148539,0.421168)
(0.051490,0.376851)
(0.047490,0.222681)
(0.051165,0.231409)
(0.045344,0.253346)
(0.103024,0.299340)
(0.036142,0.302137)
(0.197329,0.294283)
(0.048258,0.205647)
(0.041277,0.132698)
(0.146034,0.429590)
(0.041370,0.121358)
(0.159883,0.451968)
(0.064628,0.253818)
(0.189496,0.260690)
(0.130039,0.325246)
(0.060678,0.358854)
(0.212989,0.451780)
(0.100839,0.193802)
(0.177343,0.515975)
(0.067885,0.295493)
(0.064803,0.347662)
(0.083711,0.308828)
(0.135516,0.476915)
(0.035956,0.247577)
(0.041030,0.091710)
(0.038525,0.329267)
(0.038266,0.246676)
(0.131154,0.437012)
(0.051488,0.242382)
(0.050828,0.249596)
(0.074998,0.373032)
(0.174208,0.461750)
(0.079713,0.204219)
(0.168092,0.266158)
(0.176940,0.222594)
(0.068463,0.263380)
(0.199872,0.414740)
(0.043081,0.211226)
(0.143188,0.304430)
(0.042108,0.282802)
(0.197346,0.522119)
(0.047790,0.156585)
(0.082996,0.155053)
(0.143617,0.349486)
(0.210683,0.460826)
(0.116577,0.202751)
(0.043766,0.126322)
(0.059483,0.472148)
(0.083147,0.327861)
(0.190043,0.560480)
(0.203571,0.486437)
(0.042753,0.219498)
(0.198739,0.514159)
(0.197220,0.481509)
(0.053208,0.277212)
(0.042302,0.296714)
(0.128999,0.309632)
(0.182938,0.279296)
(0.057311,0.260871)
(0.048939,0.326376)
(0.138195,0.584946)
(0.173298,0.561599)
(0.106178,0.102898)
(0.197120,0.213067)
(0.167454,0.261189)
(0.184356,0.474111)
(0.037268,0.230988)
(0.044310,0.303556)
(0.042457,0.130198)
(0.044004,0.097269)
(0.175594,0.332332)
(0.159468,0.345951)
(0.164356,0.436160)
};
\addlegendentry{Hyperparameter config.}

\addplot[
    color=red,
    line width=1.2pt,
] coordinates {
(0.035956,0.221049) (0.212989,0.437657)
};
\addlegendentry{Fitted trend line}

\node[
    draw=gray,
    fill=white,
    rounded corners,
    font=\small,
    anchor=south east
] at (rel axis cs:0.95,0.05) {Pearson's $\rho$ = 0.602};

\end{axis}
\end{tikzpicture}

%% file: latex/methodology.tex
\section{\tuner~Designs}

\begin{figure}[t!]
    \centering
    \includegraphics[width=\linewidth]{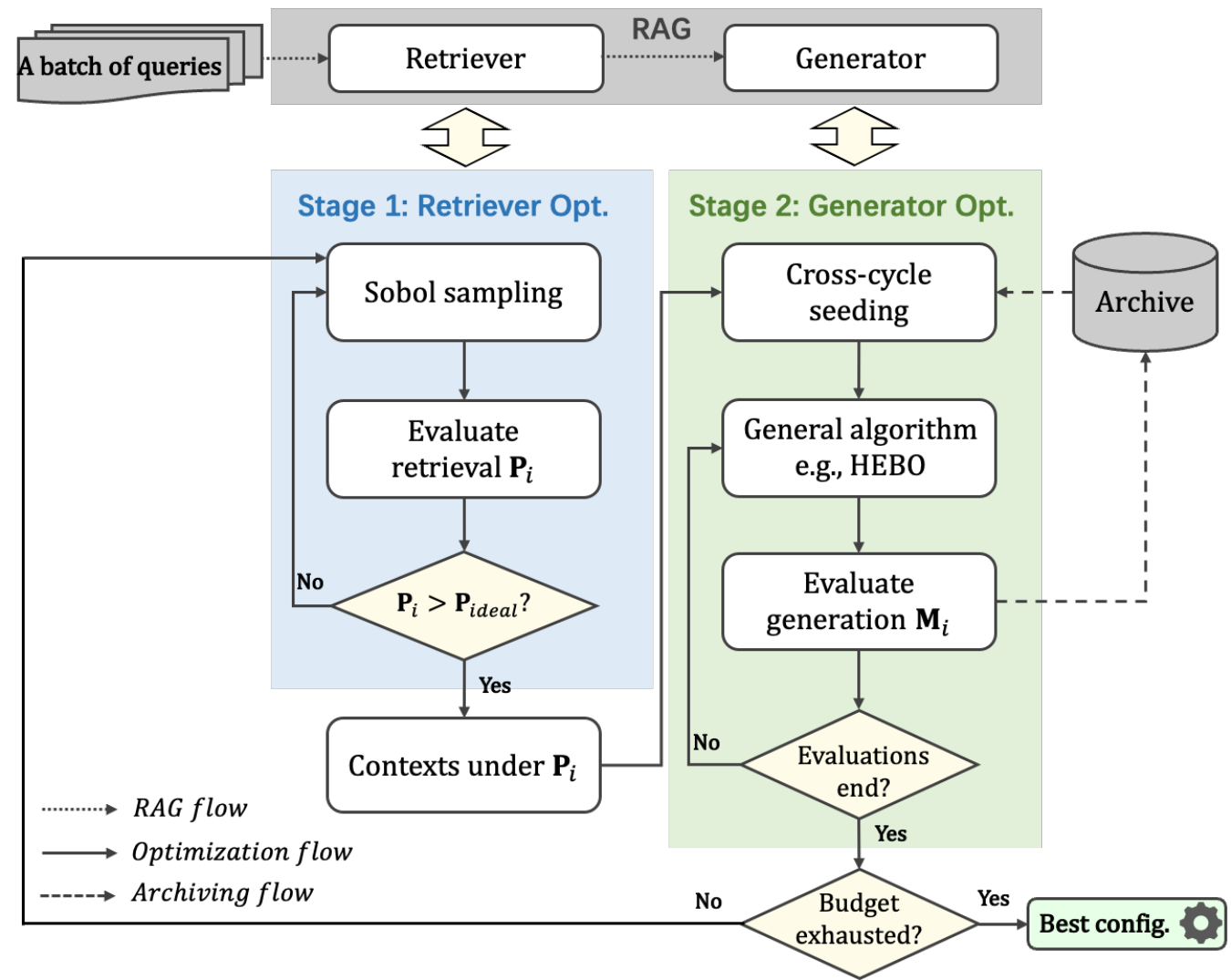}
    \caption{Overview of \tuner~to optimize RAG hyperparameters.}
    \label{fig:ragcycle_workflow}
\end{figure}

\tuner~is precisely designed to address the reformulated problem of cyclic dual-sequential hyperparameter optimization using some queries to RAG, as shown in Figure~\ref{fig:ragcycle_workflow} and Algorithm~\ref{alg:ragcycle}. For that, it has several main mechanisms:

\begin{itemize}
    \item \textbf{Cyclic Dual-Sequential Optimization:} Given a batch of queries, \tuner~cyclically optimizes the hyperparameters of the retriever (by Sobol sampling~\cite{SOBOL196786}) and generator (by any general algorithm, e.g., \texttt{HEBO} or \texttt{TPE}) in turn (lines 7-11 and 20-23). 
    


     \item \textbf{Within-Cycle Budget Provisioning:} To determine how the optimization of retriever and generator share the budget, \tuner~uses an adaptive mechanism to decide when to stop optimizing the retriever (lines 6 and 13), while for the generator, we set a cap of evaluations (line 20), reducing the costly full retrieval-generation evaluations.

     \item \textbf{Cross-Cycle Generator Seeding} To expedite the expensive optimization for the generator with a warm-start, \tuner~selects high-quality ``seeds'' of the hyperparameter configurations for generators archived across the cycles to be reused under the current cycle (lines 15-19). 
\end{itemize}

The above naturally enable \tuner~to be a general framework, such that the underlying algorithm for optimizing generator can be replaced in a plug-and-play manner.


\begin{algorithm}[t]
\caption{Pseudo code of \tuner}
\label{alg:ragcycle}
\footnotesize
\begin{algorithmic}[1]
      \State \textbf{Input:} Budget $\mathcal{B}$; generator cap $N$; archive of seeds $\mathcal{A}$; retriever quality $\mathbf{P}$; generator quality $\mathbf{M}$; corpus $\mathcal{D}$; given query batch $\mathcal{Q}$; retrieval ground truth $\mathcal{R}^*$; generation ground truth $\mathcal{G}^*$
    \State \textbf{Output:} Best hyperparameter configuration $(\boldsymbol{\phi}^*, \boldsymbol{\theta}^*)$

 \While{$\mathcal{B}$ has not been exhausted}
 \State $t=1$
 \State $\boldsymbol{\phi}_t^*=0$; $\mathcal{C}^*_t=\emptyset$
 \Statex \(\triangleright\) \textcolor{blue}{Optimizing retriever}
      \While{$\mathbf{P}_i$ of $\boldsymbol{\phi}_i < \mathbf{P}_{ideal}$}
       \State $\boldsymbol{\phi}_i \leftarrow$ find the next retriever configuration using Sobol sampling
       \State Evaluate the $\mathbf{P}_i$ of $\boldsymbol{\phi}_i$ via $\mathcal{D}$, $\mathcal{Q}$, $\mathcal{R}^*$, Equations~\ref{eq:retrieve-1}
        \State $\mathcal{C}_i \leftarrow$ retrieve contexts with $\boldsymbol{\phi}_i$. 
        \If{$\mathbf{P}_i$ of $\boldsymbol{\phi}_i>$ the $\mathbf{P}$ of $\boldsymbol{\phi}_t^*$}
             \State $\boldsymbol{\phi}_t^*=\boldsymbol{\phi}_i$; $\mathcal{C}^*_t=\mathcal{C}_i$
       \EndIf

        \Statex \(\triangleright\) \textcolor{blue}{Provisioning budget for retriever}
       \State Update $\mathbf{P}_{ideal}$ using Equations~\ref{eq:budget-1} and~\ref{eq:budget-2}
       \EndWhile

       \Statex \(\triangleright\) \textcolor{blue}{Seeding the optimization of generator}
      
      \State $\mathcal{A}'\leftarrow$ De-duplicating $\mathcal{A}$
       \State $\mathcal{H}=\emptyset$
      \State $\mathcal{H} \leftarrow$ $N\over 4$ generator configurations with top $50\%$ $\mathbf{M}$ from $\mathcal{A}'$

     \State $\mathcal{H} = \mathcal{H} \cup$ randomly generate $N\over 4$ generator configurations
     
     \State Re-evaluate the $\mathbf{M}$ of generator configurations in $\mathcal{H}$ with $\mathcal{C}^*_i$


        \Statex \(\triangleright\) \textcolor{blue}{Optimizing generator with provisioned budget}
      \State $\mathcal{H}= \mathcal{H} \cup \leftarrow$ run, e.g., \texttt{HEBO}, with $\mathcal{C}^*_i$, $\mathcal{Q}$, $\mathcal{G}^*$, and $N\over2$
       \If{$\mathbf{M}_i$ of $\forall\boldsymbol{\theta}_i \in \mathcal{H}>$  the $\mathbf{M}$ of $\boldsymbol{\theta}^*$}
            \State $\boldsymbol{\theta}^*=\boldsymbol{\theta}_i$; $\boldsymbol{\phi}^*=\boldsymbol{\phi}^*_t$
       \EndIf
       \State $\mathcal{A} = \mathcal{A} \cup \mathcal{H}$; $t=t+1$
 \EndWhile
  \State \textbf{Return} $(\boldsymbol{\phi}^*,\boldsymbol{\theta}^*)$

\end{algorithmic}
\end{algorithm}


\subsection{Dual-Sequential Optimization in Cycles}


Optimization in \tuner~occurs in two sequential stages with multiple cycles, for which we provide the details below.

\paragraph{Retriever Optimization.}

Recall from Equation~\ref{eq:stage3}, the goal of optimizing retriever is to find its hyperparameter configuration with the best $\mathbf{P}$ value, for which we define as the average similarity to the ground truth across the top $K$ documents retrieved from the corpus $\mathcal{D}$ for the given query batch $\mathcal{Q}$:
\begin{equation}
    \mathbf{P} = \frac{1}{K} \sum_{k=1}^{K} \sum_{q\in\mathcal{Q}}^{|\mathcal{Q}|}   \frac{|\mathcal{W}_{d_{q,k}} \cap \mathcal{W}_{r^*_q}|}{|\mathcal{W}_{d_{q,k}}|}
    \label{eq:retrieve-1}
\end{equation}
Here, we use lexical precision to measure the similarity between a retrieved document $d_{q,k}$ and the ground truth of context $r^*_q$ for a query $q$, where $\mathcal{W}$ denotes the unique tokens in the corresponding text. The lexical precision is chosen because it is a more efficient signal for finding useful configurations for the generator later on. Importantly, we prefer such over other semantic metrics as it is more strict, biased to prevent false positives, which is more critical for optimizing the generator.
To solve Equation~\ref{eq:stage3}, we use Sobol sampling \cite{SOBOL196786} in \tuner. This is because, since optimizing retriever only implicitly influences the final generation outcome, we seek to find its diverse yet promising hyperparameter configurations, hence increasing the likelihood to jump out of local optima when optimizing the generator later on. To that end, the Sobol sampling works as a quasi-random low-discrepancy sequence to outperform uniform sampling in high-dimensional cases. Importantly, it provides a distribution-aware exploration and stronger convergence/exploitation than random search. This is crucial for providing the subsequent generation stage with a robust variety of high-quality retrieval foundations. To adopt Sobol sampling to cope with the mixed hyperparameter for retriever, we apply randomized digital scrambling and density-preserving quantization to the Sobol sequences~\cite{Cowen-Rivers2022-HEBO}, ensuring that the low-discrepancy benefits are preserved across the mixed-integer boundary.


The $\boldsymbol{\phi}^*_t$ , which is the best hyperparameter configuration of the retriever found at cycle $t$, evaluated under the retrieval quality $\mathbf{P}$, would be fixed, and its retrieved contexts $\mathcal{C}^*_t$ would be applied in the subsequent generator optimization.

\paragraph{Generator Optimization with Fixed Retrieved Contexts.}

At cycle $t$, unlike optimizing the retriever hyperparameters, which focuses on exploration, here we need better exploitation for Equation~\ref{eq:stage2} as the evaluation of the generator is much more expensive. \tuner~can be seamlessly paired with any general algorithm, e.g., Bayesian optimization, to optimize the generation quality of RAG using the given queries, e.g., the F1-score. Yet, the retriever hyperparameters are frozen to be $\boldsymbol{\phi}^*_t$ and so do the contexts $\mathcal{C}^*_t$ retrieved thereunder for query batch $\mathcal{Q}$; only the generator's hyperparameters $\boldsymbol{\theta}_i$ are perturbed. The best configuration at cycle $t$ $(\boldsymbol{\phi}^*_t,\boldsymbol{\theta}^*_t)$ is stored.

When all budgets are exhausted, the best generator configuration found and its corresponding retrieval configuration across all cycles, $(\boldsymbol{\phi}^*,\boldsymbol{\theta}^*)$, are returned. Otherwise, we repeat from the retriever optimization in the next cycle $t+1$. All explored hyperparameter configurations of the generators and their generation quality $(\boldsymbol{\theta}_i,\mathbf{M}_i)$ are archived for the possible cross-cycle seeding next.

\subsection{Within-Cycle Budget Provisioning}


Since the nature of dual-sequential hyperparameter optimization offers fine-grained control for budget provision, we set different mechanisms for the retriever and generator.

\paragraph{Adaptive Termination for Retriever.}

For the retriever, we design an adaptive termination mechanism based on the expected progress of retrieval quality, considering all cycles so far. This makes sense since optimizing the retriever is cheap, and it often reaches an ideal retrieval plateau early, after which any more budgets spent on it would not be much beneficial.

\begin{figure}[t!]
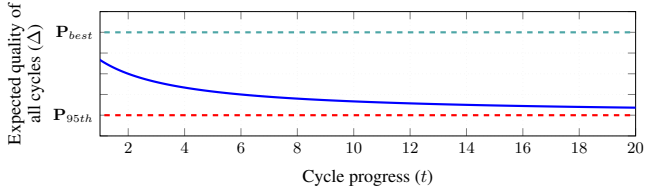

    \centering
    \includestandalone[width=\linewidth]{picture/explain_equation}
    \caption{The function of $\Delta$ changes with cycle progress $t$ ($\beta=2$).}
    \label{fig:adapt-budget}
\end{figure}

At each cycle, we compute an ideal threshold $\mathbf{P}_{ideal}$ of the obtained retrieval quality over query batch $\mathcal{Q}$:
\begin{equation}
    \mathbf{P}_{ideal} = \alpha \cdot median(\mathbf{P}_1,\mathbf{P}_2,\cdots,\mathbf{P}_n) + (1 - \alpha) \cdot  \Delta
    \label{eq:budget-1}
\end{equation}
where $(\mathbf{P}_1,\mathbf{P}_2,\cdots,\mathbf{P}_n)$ denote the locally achieved values of retrieval quality for all $n$ explored hyperparameter configurations of the retriever at the current cycle. $\Delta$ denotes the globally expected quality considering the current and all previous cycles. $\alpha$ controls the relative contribution, and we set equal weights $\alpha=0.5$ by default. $\Delta$ is formulated as:
\begin{equation}
    \Delta = \mathbf{P}_{best}\cdot(1-\frac{t}{t + \beta}) + \mathbf{P}_{95th}\cdot \frac{t}{t + \beta}
        \label{eq:budget-2}
\end{equation}
whereby $\mathbf{P}_{95th}$ and $\mathbf{P}_{best}$ denote the $95$th percentile and the best value of all retrieval quality found so far, respectively. $t$ denotes the $t$th cycle; $\frac{t}{t + \beta}$ is the function that produces a cycle-progress aware weight; we found that $\beta=2$ is appropriate. As from Figure~\ref{fig:adapt-budget}, essentially, $\mathbf{P}_{95th}$ is a more relaxed value compared with $\mathbf{P}_{best}$, serving as the lower bound of $\Delta$. As the cycle increases, $\frac{t}{t + \beta}$ would become closer to $1$, hence $\Delta$ would be closer to $\mathbf{P}_{95th}$. In this way, we can reduce the contribution of $\Delta$ for $\mathbf{P}_{ideal}$ at the latter stage of the optimization, since the results of the current cycle would gradually become more important as the optimization converges.

Finally, if a newly explored configuration of the retriever has a retrieval quality greater than or equal to $\mathbf{P}_{ideal}$, then we terminate the retriever optimization at the current cycle.  


\paragraph{Fixed Evaluation Count for Generator.}

Since optimizing the generator aligned with our main objective, we cap its budget via a fixed number of evaluations $N$, for which we set $N=10$ in this work. This is a more intuitive way for one to directly control the trade-off between budget and RAG quality. Clearly, an $N$ that is too large would reduce the contribution from optimizing/exploring the retriever, while a too small $N$ can restrict the exploitation in optimizing the generator. The sensitivity of \tuner~to $N$ can be found in \repolink{Appendix}{app:sen}.

\subsection{Cross-Cycle Generator Seeding}

A naive method would be to treat the dual-sequential optimization between cycles as independent runs, but this ignores the optimization knowledge that has been accumulated so far. Further, we observe that the hyperparameter performance of the generator exhibits significant cross-context transferability even when the frozen hyperparameters of the retriever differ.


To capitalize on that, we reuse/seed certain hyperparameter configurations, found in previous cycles, to the generator optimization at the current cycle for a warm start. The seeds can also initialize the surrogate model in, e.g., \texttt{BO}. The steps are:


\paragraph{Step 1---Archive de-duplication:} In previous cycles, it is possible to have multiple identical hyperparameter configurations of the generator with different generation quality due to their different contexts retrieved, thus we perform a de-duplication that merges those and averages their $\mathbf{M}_i$ values. Thus, all unique hyperparameter configurations in the archive would correspond to a single generation quality $(\boldsymbol{\theta}_i,\mathbf{M}_i)$.

\paragraph{Step 2---Probabilistic seeding:} To select which hyperparameter configuration $\boldsymbol{\theta}_i$ of the generator from previous cycles to seed, we rank each configuration, $r_i$, according to its (average) generation quality. For all $J$ top $50\%$ configurations, we assign a probability to each based on reciprocal of its rank:
\begin{equation}
    p(\boldsymbol{\theta}_i) = {{r^{-1}_i} \over {\sum^J_{j=1}r^{-1}_j}}
\end{equation}
$p(\boldsymbol{\theta}_i)$ indicates the likelihood of $\boldsymbol{\theta}_i$ to be used for seeding. This ensures that better performing $\boldsymbol{\theta}_i$ would have a higher probability to seed, while there is also a chance of choosing slightly worse ones, hence ensuring diversity. 

Recall that $N$ is the cap for the evaluation counts to optimize the generator, we set the number of configurations to be seeded as $N \over 2$, within which $N \over 4$ is seeded from the archive of previous cycles, while the remaining $N \over 4$ are randomly generated configurations. This ensures that the seeds can guide the hyperparameter optimization of the generator without trapping at local optima. Note that if $N \over 4$ is not an integer, we prefer the random seeds over the archived seeds, e.g., $N=10$ means that $5$ would be the seeds, in which $2$ come from the archive.

\paragraph{Step 3---Contextual re-evaluation:} Since the generation quality of generator configurations in the archive might be evaluated with a retriever hyperparameter configuration differs from the $\boldsymbol{\phi}^*_t$ at the current cycle $t$, all seeds for the generator should be re-evaluated with contexts $\mathcal{C}^*_t$ retrieved under $\boldsymbol{\phi}^*_t$.

%% file: picture/explain_equation.tex
\begin{tikzpicture}
\begin{axis}[
    width=12cm,
    height=4cm,
    xlabel={Cycle progress ($t$)},
    xlabel style={font=\LARGE},
    ylabel={Expected quality of \\all cycles ($\Delta$)},
    ylabel style={font=\LARGE,align=center,at={(-0.02,0.5)}},
    title={},
    grid=major,
    grid style={line width=.1pt, draw=gray!30, dotted},
    legend pos=north east,
    legend style={draw=black, fill=white, font=\small},
    xmin=1, xmax=20,
    ymin=0.80, ymax=0.92,
    tick label style={font=\small},
    label style={font=\normalsize},
    title style={font=\large},
    ytick={0.80, 0.82, 0.84, 0.86, 0.88, 0.90, 0.92,0.94},
    yticklabels={ , $\mathbf{P}_{95th}$,  ,  ,  , $\mathbf{P}_{best}$,  ,   },
]

\addplot[
    blue,
    very thick,
    domain=0:20,
    samples=200,
    smooth,
] {0.90 * (1 - x/(x+2)) + 0.82 * (x/(x+2))};

\addplot[
    teal!70,
    dashed,
    very thick,
    domain=0:20,
] {0.90};

\addplot[
    red,
    dashed,
    very thick,
    domain=0:20,
] {0.82};

\end{axis}
\end{tikzpicture}

%% file: latex/experiments.tex
\section{Experiment Setup}
\label{sec:experiments}

\input{table/configuration}



\input{table/results}

\subsection{Experiment Subjects}

\paragraph{Benchmarks and Knowledge Bases.} 
As the common practices for RAG, we use diverse/widely-used benchmarks~\cite{lightrag,Rag_and_Roll,HotpotQA}:


\begin{itemize}
    \item \textbf{Agriculture and Biography \cite{lightrag}:} Two domain-specific benchmarks requiring precise entity retrieval. We send all $100$ and $180$ queries to RAG.
    \item \textbf{BioASQ \cite{BioASQ}:} Biomedical questions/answers with complex terminology. We randomly sample $100$ queries per run for sending to RAG.
    \item \textbf{HotpotQA \cite{HotpotQA}:} A multi-hop reasoning benchmark for high-fidelity retrieval. Again, we randomly sample $100$ queries per run for sending to RAG.
\end{itemize}

For each benchmark, we consolidate the contextual information of source corpora for the queries into a knowledge base, which is then merged into a unified vector database. 



\paragraph{RAG System.} Like prior work \cite{autorag-hp,RAG_for_NLP,Analysis,lightrag}, we focus on a refined backbone common to most RAG systems to ensure findings are generalizable without the confounding effects of auxiliary modules like re-rankers. The joint space $\Phi \times \Theta$ in Table~\ref{tab:search_space} has up to 12 hyperparameters---a much more complex RAG case compared with the 2-7 hyperparameters used in existing work~\cite{autorag-hp,Faster}---including discrete architectural choices and continuous options. This leads to a search space of at least $1.3\times10^{18}$, if discretized.


\paragraph{LLMs.} 
We employ \texttt{Llama-3.1-8B} \cite{llama} and \texttt{Qwen3-8B} \cite{qwen3} as the generator for RAG due to their robust instruction-following capabilities. The prompt for RAG in evaluation can be found at \repolink{Appendix}{app:llm}. 


\subsection{Compared Approaches and Metric}

\paragraph{Compared Approaches.}
We compare \tuner~against several types of hyperparameter optimization algorithms:

\begin{itemize}
\item \textbf{Baseline:} We use Random Search (\texttt{Random}) and {Greedy Search} (\texttt{Greedy}) \cite{Analysis} as the baselines.


\item \textbf{General algorithms:} We also assess the state-of-the-art general algorithm for hyperparameter optimization, including classic \texttt{BO} \cite{BO_NIPS}, \texttt{TPE} \cite{TPE}, the advanced \texttt{HEBO} \cite{Cowen-Rivers2022-HEBO}. They jointly optimize all RAG hyperparameters. 


\item \textbf{RAG-specific algorithms:} We compare two state-of-the-art algorithms designed for optimizing RAG hyperparameters: \texttt{RayTune} \cite{RayTune} and \texttt{AutoRAG-HP} \cite{autorag-hp}: the former is a tailored TPE for RAG, while the latter relies on hierarchical multi-armed bandit. Again, they do not distinguish the hyperparameters of the retriever and generator in the optimization.

\end{itemize}

Despite being relevant, we have had to omit some algorithms due to different reasons: (1) Prompt-only algorithms (e.g., \texttt{MIPROv2} \cite{Miprov2}) and single-stage approaches (e.g., \texttt{AdaptiveRAG} \cite{Adaptiverag}) are omitted, as it is unfair to compare with them and they are orthogonal to this work. (2) Multi-fidelity algorithms (e.g., \texttt{BOHB} \cite{BOHB} and \texttt{DEHB} \cite{DEHB}) are also ruled out, because multi-fidelity is an additional feature that can easily be paired with \tuner~ under its cyclic dual-sequential optimization. (3) Since this work focuses on single objective optimization, we do not consider multi-objective RAG optimization algorithms such as \texttt{METIS} \cite{Metis}. 




\paragraph{Evaluation Metric.}
To evaluate the generation quality of optimized RAG, we focus on the token-level F1-score (kindly see \repolink{Appendix}{app:metrics}) based on the ground truth answers of the given queries of a benchmark as they are sent to RAG, which is also the main optimization objective. F1-score is chosen because it is robust to data imbalance and has been widely used for RAG evaluation~\cite{Metis,SQuAD,DBLP:journals/corr/abs-2405-07437}. Since RAG follows the zero-shot learning paradigm without the need for any (re-)training on the given queries when changing hyperparameters, the above can evaluate/test the generalizability of RAG under optimized hyperparameters, resembling the common way of using a validation loss, which is both an optimization objective and evaluation metric, in classic AutoML~\cite{DBLP:journals/csur/KarmakerHSXZV22}---the same is also applied in \texttt{AutoRAG-HP}~\cite{autorag-hp}. Having said that, we have indeed additionally evaluated the optimized configurations on holdout queries from \textit{BioASQ} and \textit{HotpotQA}, which have sufficient samples (\repolink{Appendix}{app:holdout}).



\subsection{Implementation and Setting Details}

The RAG is implemented via \texttt{LangChain}\footnote{\url{https://github.com/langchain-ai/langchain}}, with \texttt{OpenBox} \cite{openbox} for \tuner~and other hyperparameter optimization algorithms, or otherwise we directly use the code provided by their authors. All experiments run on a cluster of 10 machines, each equipped with an Intel CPU of 24 cores and 32 threads, 128GB RAM, and a NVIDIA RTX 5090 GPU. We set the hyperparameters of \tuner~as the aforementioned defaults, and those of the other algorithms as what used by their authors. The details can be found in \repolink{Appendix}{app:settings}. To complete the experiment within a reasonable time while ensuring realism, for each algorithm/run, we set a budget of one hour wall-clock time (where most algorithms have already converged reasonably) and repeat $10$ independent runs. In total, the entire experiment took $\approx1,000$ CPU/GPU hours. 




%% file: table/configuration.tex
\begin{table}[t!]
\centering

\resizebox{\columnwidth}{!}{%
\begin{tabular}{@{}lllr@{}}
\toprule
\textbf{Stage} & \textbf{Hyperparameter} & \textbf{Type} & \textbf{Range / Values} \\ \midrule
\multirow{7}{*}{\textbf{Retriever ($\Phi$)}} & Database Choice & Categorical & \{DuckDB, Chroma, FAISS\} \\
 & Chunk Size & Integer & $[256, 1024]$ \\
 & Chunk Overlap & Integer & $[32, 128]$ \\
 & Embedding Temperature & Float & $[0.0, 1.0]$ \\
 & Embedding Window & Integer & $[512, 2048]$ \\
 & Embedding Repeat Penalty & Float & $[0.9, 1.5]$ \\
 & Embedding Top-$k$ & Integer & $[10, 100]$ \\ \midrule
\multirow{5}{*}{\textbf{Generator ($\Theta$)}} & Retrieval Numbers ($K$) & Integer & $[1, 10]$ \\
 & Generation Temperature & Float & $[0.0, 1.0]$ \\
 & Generation Window & Integer & $[512, 8192]$ \\
 & Generation Repeat Penalty & Float & $[0.9, 1.5]$ \\
 & Generation Top-$k$ & Integer & $[10, 100]$ \\ \bottomrule
\end{tabular}%
}
\caption{RAG system and the hyperparameter space studied.}
\label{tab:search_space}
\end{table}

%% file: table/results.tex
\begin{table*}[t!]
\centering

\begin{adjustbox}{width=\textwidth}\begin{tabular}{lllllllll}\toprule
\multirow{2}{*}{\textbf{Algorithm}} & \multicolumn{4}{c}{\textbf{Llama-3.1-8B}} & \multicolumn{4}{c}{\textbf{Qwen-3-8B}} \\ \cmidrule(lr){2-5} \cmidrule(lr){6-9}
 & \textbf{Agriculture} & \textbf{Biography} & \textbf{HotpotQA} & \textbf{BioASQ} & \textbf{Agriculture} & \textbf{Biography} & \textbf{HotpotQA} & \textbf{BioASQ}  \\ \midrule
\texttt{HEBO} & 0.380$_{\pm 0.016}$ & 0.335$_{\pm 0.011}$ & 0.571$_{\pm 0.065}$ & 0.282$_{\pm 0.010}$ & 0.296$_{\pm 0.004}$ & 0.299$_{\pm 0.003}$ & 0.622$_{\pm 0.128}$ & 0.228$_{\pm 0.002}$ \\
\tuner~(\texttt{HEBO}) & \textbf{0.383}$_{\pm \mathbf{0.013}}$ & \textbf{0.352}$_{\pm \mathbf{0.007}}$ & \textbf{0.626}$_{\pm \mathbf{0.017}}$ & \textbf{0.293}$_{\pm \mathbf{0.011}}$ & \textbf{0.305}$_{\pm \mathbf{0.009}}$ & \textbf{0.317}$_{\pm \mathbf{0.007}}$ & \textbf{0.759}$_{\pm \mathbf{0.036}}$ & \textbf{0.235}$_{\pm \mathbf{0.004}}$ \\ \hline
\texttt{BO} & 0.324$_{\pm 0.025}$ & 0.322$_{\pm 0.019}$ & 0.550$_{\pm 0.103}$ & \textbf{0.266}$_{\pm \mathbf{0.016}}$ & 0.285$_{\pm 0.011}$ & 0.303$_{\pm 0.009}$ & 0.475$_{\pm 0.141}$ & \textbf{0.236}$_{\pm \mathbf{0.003}}$ \\
\tuner~(\texttt{BO}) & \textbf{0.366}$_{\pm \mathbf{0.019}}$ & \textbf{0.343}$_{\pm \mathbf{0.014}}$ & \textbf{0.612}$_{\pm \mathbf{0.043}}$ & 0.263$_{\pm 0.013}$ & \textbf{0.294}$_{\pm \mathbf{0.014}}$ & \textbf{0.309}$_{\pm \mathbf{0.009}}$ & \textbf{0.619}$_{\pm \mathbf{0.041}}$ & 0.231$_{\pm 0.006}$ \\ \hline
\texttt{TPE} & 0.359$_{\pm 0.026}$ & 0.335$_{\pm 0.011}$ & 0.548$_{\pm 0.042}$ & 0.276$_{\pm 0.012}$ & 0.286$_{\pm 0.008}$ & 0.299$_{\pm 0.004}$ & 0.639$_{\pm 0.145}$ & \textbf{0.231}$_{\pm \mathbf{0.003}}$ \\
\tuner~(\texttt{TPE}) & \textbf{0.373}$_{\pm \mathbf{0.009}}$ & \textbf{0.345}$_{\pm \mathbf{0.004}}$ & \textbf{0.597}$_{\pm \mathbf{0.014}}$ & \textbf{0.284}$_{\pm \mathbf{0.008}}$ & \textbf{0.289}$_{\pm \mathbf{0.006}}$ & \textbf{0.300}$_{\pm \mathbf{0.002}}$ & \textbf{0.734}$_{\pm \mathbf{0.037}}$ & 0.228$_{\pm 0.002}$ \\ \bottomrule
\end{tabular} \end{adjustbox}
\caption{Comparing \tuner~paired with \texttt{HEBO}, and \texttt{BO}, and \texttt{TPE} against their vanilla original versions for optimizing RAG over 10 runs. Values show mean$_{\pm \text{Std}}$ of F1-scores across the runs; the best results are highlighted in \textbf{bold}. Details of the trajectories can be found in \repolink{Appendix}{app:rq1}.}
\label{tab:multimodel_main}
\end{table*}

%% file: latex/results.tex
\section{Evaluation Results}
\label{sec:results}


\begin{figure}[t!]
    \centering
    \includegraphics[width=\linewidth]{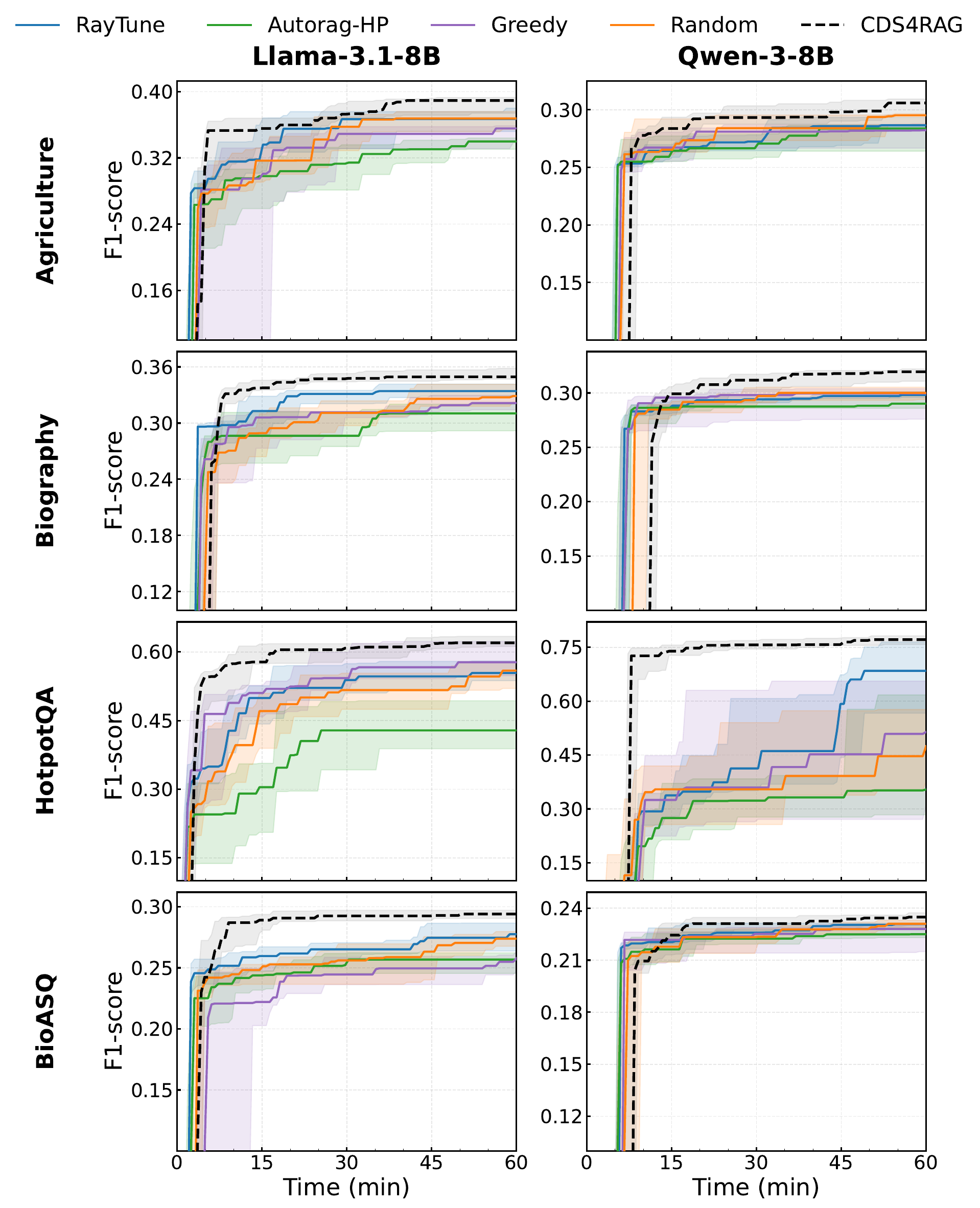}
    \caption{Optimization trajectories of \tuner~against the state-of-the-art algorithms for RAG over 10 runs}
    \label{fig:rq2}
\end{figure}

\subsection{Benefits of Cyclic Dual-Sequential Optimization}
To assess the benefits of cyclic dual-sequential optimization, we pair \tuner~with \texttt{HEBO}, \texttt{BO}, and \texttt{TPE} while comparing them against their respective single-stage optimization (jointly optimizing retriever and generator). As shown in Table~\ref{tab:multimodel_main}, \tuner~yields considerably better results than its counterparts in 21/24 cases. For both LLMs, the improvement of \tuner~appears to be more significant on \textit{HotpotQA} than on others, due to the queries therein being more complex. This makes sense, since if the queries in a benchmark are easy, then the room to be improved is limited. We see that \tuner~paired with \texttt{HEBO} has the best result than other variants, since \texttt{HEBO} perform better than \texttt{BO} and \texttt{TPE} in general. On holdout queries, \tuner~also lead to significant improvements overall (see \repolink{Appendix}{app:holdout}). All those confirm the benefits of distinguishing the hyperparameters of retriever and generator in \tuner~and its algorithm-agnostic capability.




\subsection{Comparison with State-of-the-Art Algorithms}

We further evaluate \tuner~paired with \texttt{HEBO} against four baselines/state-of-the-art algorithms. As shown in Figure~\ref{fig:rq2}, remarkably, \tuner~achieves the significantly better results on all cases. Compared with \texttt{RayTune}, \tuner~shows a clear advantage, reaching 0.759 F1-score on \textit{HotpotQA} for \texttt{Qwen-3-8B} versus the 0.639 of \texttt{RayTune}. The superiority over RAG-specific algorithms like \texttt{Autorag-HP} is even more pronounced; for example, on \textit{BioASQ} with \texttt{Llama-3.1-8B}, \tuner~attains 0.293 compared with 0.259; on \textit{HotpotQA} with \texttt{Qwen-3-8B}, \tuner~improves \texttt{Autorag-HP} by $1.54\times$. Most importantly, after the early phase of retriever optimization, the trajectory of \tuner~improves much steeper than those of the others with a faster and better convergence/speedup in a matter of a few minutes, showing its superiority even under a limited budget. On holdout queries, the conclusion remains unchanged: over all LLMs/datasets, \tuner~(with \texttt{HEBO}) achieves 0.4180 F1-score compared with the 0.3306--0.3853 of the other state-of-the-art algorithms ($10\%$--$25\%$ better); see \repolink{Appendix}{app:holdout}.




\input{table/results_ablation}
\subsection{Ablation Analysis}

To confirm the contributions of individual design in \tuner, we replace the within-cycle budget provisioning with fixed, equally dividing budgets to retriever and generator, denoted as \texttt{w/o pro}; we also remove the cross-cycle generator seeding, denoted as \texttt{w/o seeding}. As shown in Table~\ref{tab:rq3_ablation}, we see that removing either of the above mechanisms is harmful in general. Notably, interactions exist between the two mechanisms: the budget provision determines the number of possible cycles, which then influences the cross-cycle seeding that impacts the convergence speed of RAG hyperparameter optimization.

\begin{figure}[t!]
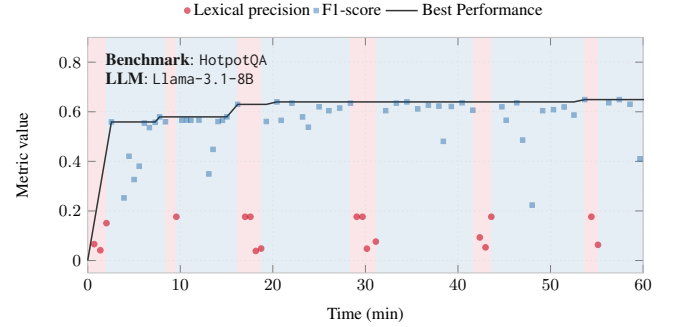

    \centering
    \includestandalone[width=\linewidth]{picture/case_study}
    \caption{A case of \tuner~for RAG that completes $T=6$ cycles.}
    \label{fig:case}
\end{figure}

\subsection{A Case Study}
For the case in Figure~\ref{fig:case}, \tuner~completes 6 cycles with $\approx17\%$ budgets spent on retrieval evaluations only. In less than 10 minutes, it obtains an excellent generation quality already, followed by F1-score increases at a slow, but steady pace: the marginal improvements of the overall lexical precision for the retriever always help to yield better/more promising overall F1-score for the generator---well-aligned with our hypothesis. 




%% file: table/results_ablation.tex
\begin{table}[t!]
\centering

\begin{adjustbox}{width=\linewidth}
\begin{tabular}{l llll}
\toprule
\textbf{Algorithm} & \textbf{Agriculture} & \textbf{Biography} & \textbf{HotpotQA} & \textbf{BioASQ}  \\ \midrule
\multicolumn{5}{l}{\cellcolor{gray!30}\textbf{Llama-3.1-8B}} \\
\texttt{w/o pro} & \textbf{0.387}$_{\pm \mathbf{0.009}}$ & 0.348$_{\pm0.009}$ & 0.621$_{\pm0.009}$ & 0.289$_{\pm0.012}$ \\
\texttt{w/o seeding} & 0.375$_{\pm0.009}$ & 0.347$_{\pm0.008}$ & 0.593$_{\pm0.021}$ & 0.278$_{\pm0.007}$ \\
\tuner & 0.383$_{\pm0.013}$ & \textbf{0.352}$_{\pm \mathbf{0.007}}$ & \textbf{0.626}$_{\pm \mathbf{0.017}}$ & \textbf{0.293}$_{\pm \mathbf{0.011}}$ \\
\midrule
\multicolumn{5}{l}{\cellcolor{gray!30}\textbf{Qwen-3-8B}} \\
 \texttt{w/o pro} & 0.291$_{\pm0.009}$ & 0.299$_{\pm0.003}$ & 0.727$_{\pm0.053}$ & 0.229$_{\pm0.003}$ \\
\texttt{w/o seeding} & 0.289$_{\pm0.006}$ & 0.300$_{\pm0.004}$ & 0.752$_{\pm0.034}$ & 0.230$_{\pm0.003}$ \\
\tuner & \textbf{0.305}$_{\pm \mathbf{0.009}}$ & \textbf{0.317}$_{\pm \mathbf{0.007}}$ & \textbf{0.759}$_{\pm \mathbf{0.036}}$ & \textbf{0.235}$_{\pm \mathbf{0.004}}$ \\
\bottomrule
\end{tabular}
\end{adjustbox}
\caption{Ablation results on \tuner~over 10 runs. The formate is the same as Table~\ref{tab:multimodel_main}. Details of the trajectories can be found in \repolink{Appendix}{app:rq3}.}
\label{tab:rq3_ablation}
\end{table}

%% file: picture/case_study.tex
\begin{tikzpicture}
\begin{axis}[
    width=12cm, 
    height=6cm,
    xlabel={Time (min)},
    ylabel={Metric value},
    label style={font=\small},
    tick label style={font=\footnotesize},
    xmin=0, xmax=60,
    ymin=-0.05, ymax=0.9,
    x filter/.code={\pgfmathparse{#1/60}\pgfmathresult},
    grid=major,
    grid style={dotted, gray!30},
    axis line style={gray!50},
    legend style={
        at={(0.5, 1.05)}, 
        anchor=south, 
        legend columns=-1,
        draw=none, 
        fill=none,
        font=\footnotesize
    },
    axis on top=false,
]

\begin{scope}[on background layer]
    \fill[rwth-red!10] (axis cs:0/60,-0.05) rectangle (axis cs:121.72/60,0.9);
    \fill[rwth-blue!10] (axis cs:121.72/60,-0.05) rectangle (axis cs:504.16/60,0.9);
    \fill[rwth-red!10] (axis cs:504.16/60,-0.05) rectangle (axis cs:574.85/60,0.9);
    \fill[rwth-blue!10] (axis cs:574.85/60,-0.05) rectangle (axis cs:972.04/60,0.9);
    \fill[rwth-red!10] (axis cs:972.04/60,-0.05) rectangle (axis cs:1124.51/60,0.9);
    \fill[rwth-blue!10] (axis cs:1124.51/60,-0.05) rectangle (axis cs:1703.26/60,0.9);
    \fill[rwth-red!10] (axis cs:1703.26/60,-0.05) rectangle (axis cs:1867.78/60,0.9);
    \fill[rwth-blue!10] (axis cs:1867.78/60,-0.05) rectangle (axis cs:2496.22/60,0.9);
    \fill[rwth-red!10] (axis cs:2496.22/60,-0.05) rectangle (axis cs:2616.24/60,0.9);
    \fill[rwth-blue!10] (axis cs:2616.24/60,-0.05) rectangle (axis cs:3221.83/60,0.9);
    \fill[rwth-red!10] (axis cs:3221.83/60,-0.05) rectangle (axis cs:3307.53/60,0.9);
    \fill[rwth-blue!10] (axis cs:3307.53/60,-0.05) rectangle (axis cs:3600.00/60,0.9);
\end{scope}

\addplot[
    only marks,
    mark=*,
    mark size=1.5pt,
    mark options={fill=rwth-red, draw=none, opacity=0.6},
    color=rwth-red
] coordinates {
(42.22,0.0661) (81.86,0.0414) (121.72,0.1508) (574.85,0.1765) (1020.54,0.1765) (1054.85,0.1766) (1089.78,0.0386) (1124.51,0.0484) (1743.84,0.1766) (1781.18,0.1766) (1809.80,0.0479) (1867.78,0.0764) (2541.67,0.0932) (2578.66,0.0532) (2616.24,0.1766) (3264.77,0.1766) (3307.53,0.0634) 
};
\addlegendentry{Lexical precision}

\addplot[
    only marks,
    mark=square*,
    mark size=1.2pt,
    mark options={fill=rwth-blue, draw=none, opacity=0.4},
    color=rwth-blue
] coordinates {
(153.79,0.5589) (235.21,0.2526) (267.77,0.4206) (300.84,0.3266) (334.09,0.3805) (367.03,0.5546) (399.87,0.5362) (437.66,0.5576) (466.34,0.5794) (504.16,0.5597) (609.72,0.5662) (641.10,0.5667) (668.83,0.5657) (721.10,0.5666) (785.22,0.3493) (812.51,0.4484) (845.14,0.5607) (873.59,0.5653) (902.47,0.5797) (972.04,0.6298) (1157.80,0.5607) (1226.49,0.6398) (1254.01,0.5653) (1323.71,0.6348) (1392.67,0.5792) (1430.25,0.5378) (1500.03,0.6200) (1563.31,0.6043) (1633.32,0.6149) (1703.26,0.6347) (1931.50,0.6043) (2000.60,0.6347) (2069.21,0.6398) (2139.01,0.6117) (2208.17,0.6268) (2276.30,0.6229) (2305.23,0.4807) (2357.04,0.6213) (2426.62,0.6363) (2496.22,0.6064) (2684.77,0.6200) (2712.52,0.5657) (2781.20,0.6363) (2819.19,0.4858) (2880.83,0.2235) (2948.99,0.6045) (3019.38,0.6085) (3088.36,0.6194) (3151.96,0.5868) (3221.83,0.6491) (3377.61,0.6363) (3446.16,0.6491) (3515.21,0.6304) (3580.09,0.4104) (3650.17,0.2694) 
};
\addlegendentry{F1-score}

\addplot[
    thick,
    color=black!80
] coordinates {
(0,0) (153.79,0.5589) (235.21,0.5589) (267.77,0.5589) (300.84,0.5589) (334.09,0.5589) (367.03,0.5589) (399.87,0.5589) (437.66,0.5589) (466.34,0.5794) (504.16,0.5794) (609.72,0.5794) (641.10,0.5794) (668.83,0.5794) (721.10,0.5794) (785.22,0.5794) (812.51,0.5794) (845.14,0.5794) (873.59,0.5794) (902.47,0.5797) (972.04,0.6298) (1157.80,0.6298) (1226.49,0.6398) (1254.01,0.6398) (1323.71,0.6398) (1392.67,0.6398) (1430.25,0.6398) (1500.03,0.6398) (1563.31,0.6398) (1633.32,0.6398) (1703.26,0.6398) (1931.50,0.6398) (2000.60,0.6398) (2069.21,0.6398) (2139.01,0.6398) (2208.17,0.6398) (2276.30,0.6398) (2305.23,0.6398) (2357.04,0.6398) (2426.62,0.6398) (2496.22,0.6398) (2684.77,0.6398) (2712.52,0.6398) (2781.20,0.6398) (2819.19,0.6398) (2880.83,0.6398) (2948.99,0.6398) (3019.38,0.6398) (3088.36,0.6398) (3151.96,0.6398) (3221.83,0.6491) (3377.61,0.6491) (3446.16,0.6491) (3515.21,0.6491) (3580.09,0.6491) (3650.17,0.6491) 
};
\addlegendentry{Best Performance}

\node[
    fill opacity=0.9,     
    rounded corners,      
    font=\small,   
    align=left,           
    anchor=north west     
] at (rel axis cs:0.02,0.96) { 
    \textbf{Benchmark}: \texttt{HotpotQA}\\
    \textbf{LLM}: \texttt{Llama-3.1-8B}
};

\end{axis}
\end{tikzpicture}

%% file: latex/conclusion.tex
\section{Conclusion}

This paper presents \tuner, a framework that uniquely distinguishes the hyperparameters of the retriever and generator, optimizing full hyperparameters in a sequential and cyclic manner. Such a paradigm is agnostic to the underlying general algorithm for the generator, enabling finer-grained budget provision within each cycle and cross-cycle generator seeding that can expedite the optimization via reusing promising hyperparameter configurations. Experiments on four benchmarks and two LLMs show that \tuner~can boost vanilla algorithms in general while outperforming state-of-the-art algorithms in all cases with up to $1,54\times$ improvement and better speedup. In future work, we seek to extend \tuner~with more optional stages of RAG and consider multiple optimization objectives.



%% file: appendix-content.tex
\appendix

\section{More Correlations Between Retrieval and Generation Quality}
\label{app:cases}

\begin{figure}[t]
    \centering
    \includegraphics[width=\columnwidth]{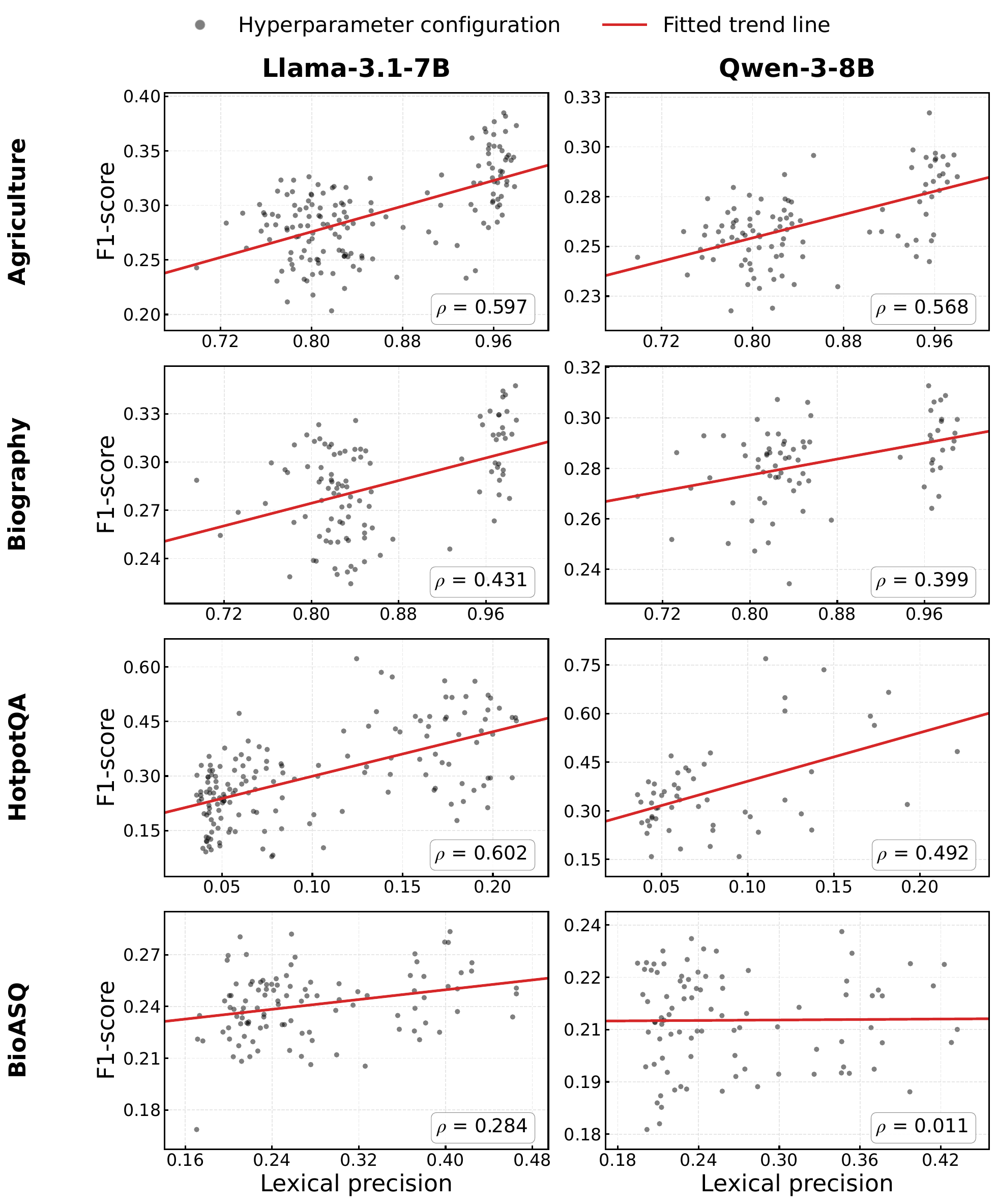}
    \caption{Full correlation analysis between retriever and generator quality for RAG over all cases studied (an extension of the Figure 2 in the main paper). The Pearson correlation is denoted by $\rho$.}
    \label{fig:full-corr}
\end{figure}

Figure \ref{fig:full-corr} presents the Pearson correlations between the quality of retrieval and generation over different RAG hyperparameter configurations using given queries on all cases studied. This extends the Figure 2 from the main paper. We see that there exist a good extent of positively monotonic correlation in general (with mostly moderate to strong signal), which enables \tuner~to yield considerable advances over the state-of-the-art algorithms, thanks to the concept of cyclic dual-sequential hyperparameter optimization. The only exception is that RAG with \texttt{Qwen3-8B} exhibits a relatively unclear correlation on the \texttt{BioASQ} benchmark, which is consistent with our experimental results, as in that case, the superiority of \tuner~is more blurred than in the other LLMs/benchmarks. 


\section{Additional Experiment Setup Details}

\subsection{RAG Prompt Template}
\label{app:llm}

To ensure a fair comparison, we employ a unified prompt template across all algorithms. As can be seen from Figure~\ref{fig:prompt_template}, the prompt consists of a fixed system instruction and a user input that incorporates the retrieved documents and the query. The content highlighted by $\{\cdots\}$ would be replaced by the actual query details and contexts retrieved.

\begin{figure*}[h]
    \centering
    \begin{tcolorbox}[
        colback=gray!5, 
        colframe=black, 
        title=\textbf{Query Template}, 
        fonttitle=\bfseries,
        boxrule=0.8pt,
        sharp corners=south, 
    ]
    \textbf{[System Message]} \\
    You are a helpful, respectful and honest assistant.
    Always answer as helpfully as possible, while being safe.
    Your answers should not include any harmful, unethical, racist, sexist, toxic, dangerous, or illegal content.
    Please ensure that your responses are socially unbiased and positive in nature.
    If you need include the answer just output the answer, no need to explain.
    
    \vspace{0.2cm}
    \hrule
    \vspace{0.2cm}
    
    \textbf{[User Message]} \\
    \texttt{[document]: \{retrieved context\}} \\
    \texttt{[conversation]: \{question\}.} 
    Answer solely based on the provided context. If information is absent, state `I do not know'.
    \end{tcolorbox}
    \caption{The unified prompt template for RAG.}
    \label{fig:prompt_template}
\end{figure*}

\begin{table*}[t]
    \centering
  
    \resizebox{\textwidth}{!}{%
    \begin{tabular}{lll}
        \toprule
        \textbf{Method} & \textbf{Hyperparameter Values} & \textbf{Description} \\
        \midrule
        \texttt{HEBO} & Model = GP; Acq = MACE; Opt = NSGA-II; Initial size = $5$ & Heteroscedastic evolutionary BO using MACE acquisition function. \\
        \texttt{BO} & Model = GP; Acq = EI; Initial = Max-min; Initial size = $5$ & Standard Bayesian Optimization with Gaussian Process surrogate. \\
        \texttt{TPE} & Initial size = $5$; $\gamma = 0.25$; & Use kernel density estimation to distinguish good and bad samples. \\
        \texttt{RayTune} & Model = TPE; Initial size = $5$; $\gamma = 0.25$; & Tailor Tree-structured Parzen Estimator with default settings. \\
        \texttt{AutoRAG-HP} & Space = Discrete; Quantization levels = $5$ & Use Hierarchical Multi-Armed Bandit on discretized parameters. \\
        \texttt{Greedy} & Space = Discrete; Quantization levels = $5$   & Optimize iteratively over a finite candidate set. \\
        \texttt{Random} & N/A & Performs uniform sampling with distinct seeds. \\
        \tuner & $\alpha=0.5$; $\beta=2$; $N=10$ & Optimize via a new formulation of cyclic dual-sequential problem.\\
        \bottomrule
    \end{tabular}%
    }
      \caption{Detailed specifications and configurations of the baseline algorithms used in the experiments.}
    \label{tab:baseline_specs}
\end{table*}

\subsection{Evaluation Metric Details}
\label{app:metrics}

We evaluate the generation quality using the token-level F1-score, which measures the overlap between the RAG generations and the ground truth answers. For a given pair consisting of the generation $y$ from a query and a reference $r$, let $T(y)$ and $T(r)$ denote the multisets of tokens derived from them. Specifically, for RAG with \texttt{Qwen-3-8B}, we explicitly exclude the generated reasoning process (chain-of-thought) from $y$ and only calculate the score based on the final answer. The number of matched tokens is calculated based on the multiset intersection:

\begin{equation}
    N_{\text{match}} = \sum_{w \in V} \min(\text{count}(w, T(y)),  \text{count}(w, T(r)))
\end{equation}

\noindent where $V$ is the vocabulary of unique tokens, and $\text{count}(w, S)$ is the frequency of token $w$ in multiset $S$. The Precision ($P$) and Recall ($R$) for this pair are defined as:

\begin{equation}
    P = \frac{N_{\text{match}}}{|T(y)|}, \quad R = \frac{N_{\text{match}}}{|T(r)|}
\end{equation}

\noindent The F1-score for the individual query is the harmonic mean of precision and recall:

\begin{equation}
    \text{F1} = 2 \cdot \frac{P \cdot R}{P + R}
\end{equation}

\noindent To evaluate the overall performance on the given batch of queries $\mathcal{Q}$ consisting of $n$ queries, we calculate the average F1-score across all given queries:

\begin{equation}
    \text{F1-score} = \frac{1}{n} \sum_{i=1}^{n} \text{F1}_i
\end{equation}

\noindent where $\text{F1}_i$ denotes the F1-score calculated for the $i$-th query-generation pair in the dataset.

\subsection{Parameter Settings for Algorithms}
\label{app:settings}

We delineate the detailed configurations of the counterpart algorithms and \tuner~in Table~\ref{tab:baseline_specs}. For \texttt{Random Search}, hyperparameter configurations are sampled uniformly from the continuous search space using distinct random seeds to ensure reliability. Regarding \texttt{Greedy Search} and \texttt{AutoRAG-HP}, we adapt the experimental setting by discretizing all continuous hyperparameters. This is necessary because \texttt{AutoRAG-HP} originally operates on discrete variables, while \texttt{Greedy Search} requires a finite candidate set for iterative optimization. Consequently, a unified discrete search space is employed for these methods.

For other baselines, \texttt{RayTune} employs the Tree-structured Parzen Estimator (TPE) with a splitting threshold of $\gamma = 0.25$. \texttt{Bayesian Optimization (BO)} utilizes a Gaussian Process (GP) surrogate with Expected Improvement (EI) as the acquisition function, initialized by maximizing the internal minimum distance. \texttt{HEBO} leverages a GP surrogate with Multi-Armed Contextual Bandit Estimation (MACE) and NSGA-II as the optimizer. To maintain a fair comparison with \tuner, the number of initial points is set to 5 for all methods requiring an initialization phase.

\section{Additional Results}

Here, we provide detailed trajectories of the results per case that were omitted from the main paper due to space constraints.

\subsection{Optimization Trajectories for Comparing \tuner~against vanilla Algorithms}
\label{app:rq1}

\begin{figure*}[t]
    \centering
    \includegraphics[width=0.91\linewidth]{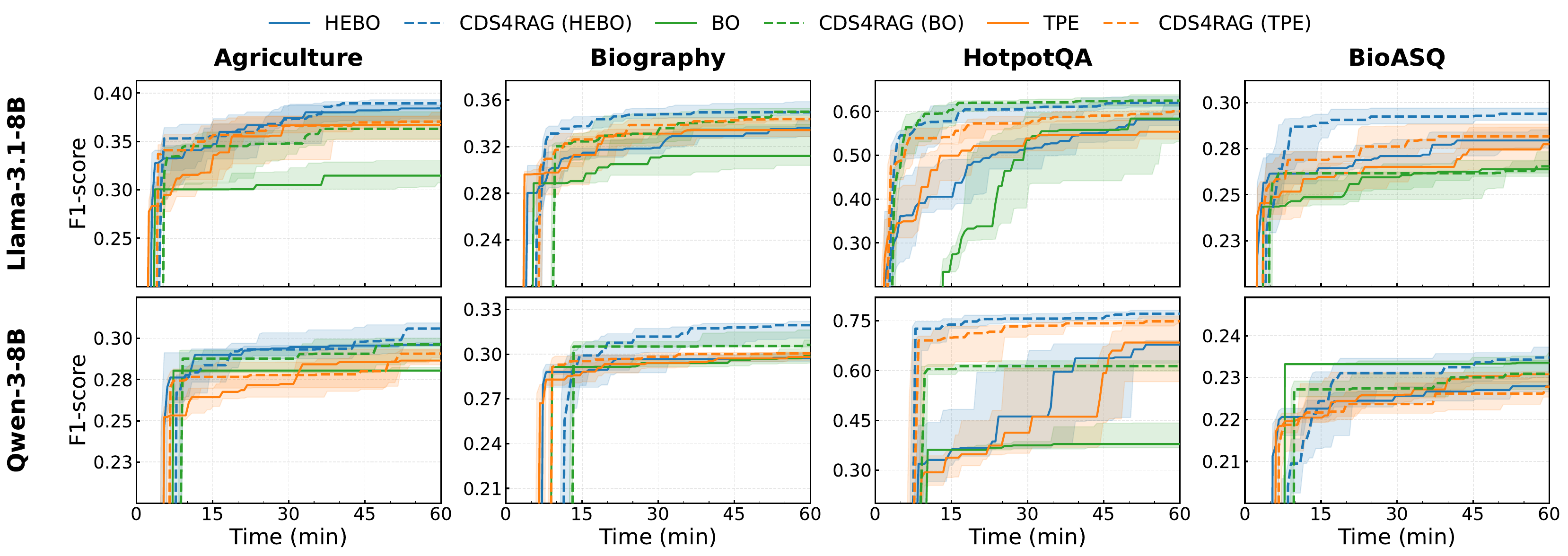}
    \caption{Comparing \tuner~paired with \texttt{HEBO}, \texttt{BO}, and \texttt{TPE} against their vanilla original versions for optimizing RAG over 10 runs.}
    \label{fig:rq1}
\end{figure*}

As in Figure~\ref{fig:rq1},
we show the optimization trajectories of the F1-score over a 60 minutes period and compare the proposed \tuner~indicated by dashed lines integrated with three SOTA optimization algorithms—\texttt{HEBO}, \texttt{BO}, and \texttt{TPE}—against the standard vanilla versions of these algorithms indicated by solid lines. Clearly, the \tuner~variants generally converge faster and achieve higher F1-score compared to their corresponding counterparts with better speedup under limited budget, hence saving the cost, particularly in the \texttt{HotpotQA} benchmark.

\subsection{Optimization Trajectories for Ablation Study}
\label{app:rq3}

\begin{figure*}[t!]
    \centering
    \includegraphics[width=0.91\linewidth]{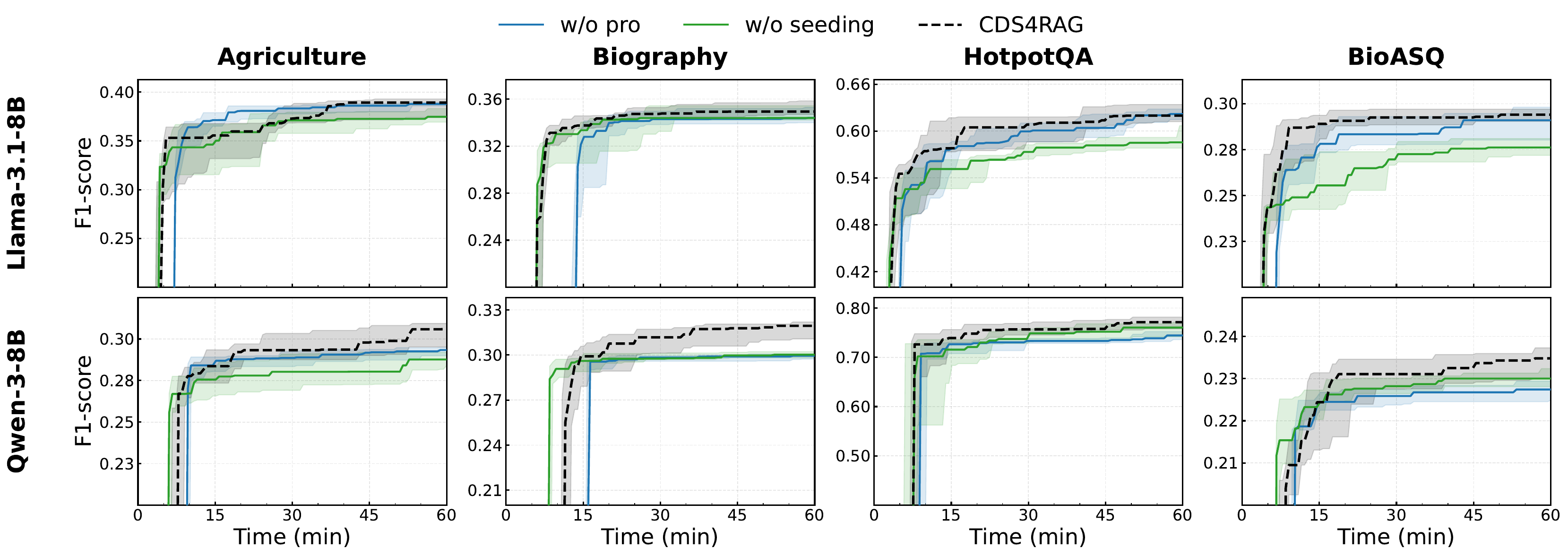}
    \caption{Ablation results on \tuner~over 10 runs.}
    \label{fig:rq3}
\end{figure*}

The trajectories for the ablation study has been shown in Figure~\ref{fig:rq3},
where we replace the within-cycle budget provisioning with fixed, equally dividing budgets to retriever and generator, denoted as \texttt{w/o pro}; we also remove the cross-cycle generator seeding, denoted as \texttt{w/o seeding}. Again, the experiments are with two LLMs across all four benchmarks with 10 runs per setting as before. Overall, the ablation results demonstrate that our \tuner~consistently enhances the performance of both models across most tasks with improved speedup and time/cost saving, even considering some limited budget scenarios.

\begin{figure*}[t!]
    \centering
    \includegraphics[width=0.91\linewidth]{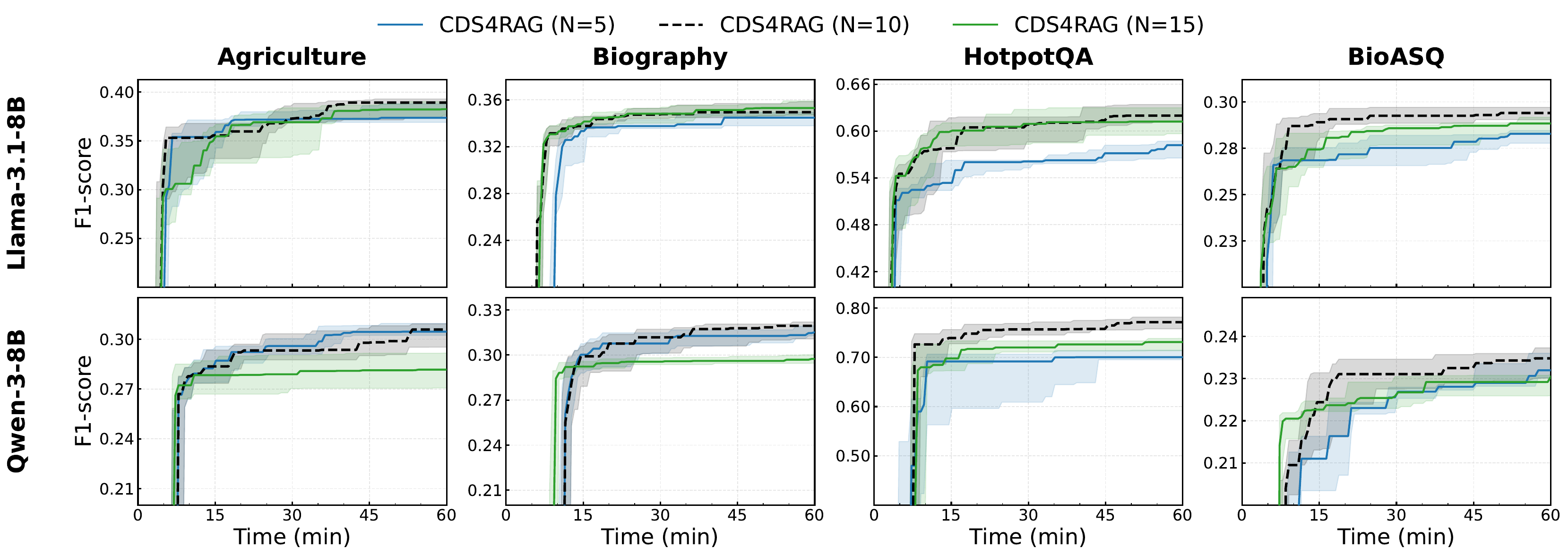}
    \caption{Sensitivity of \tuner~ to $N$ over 10 runs.}
    \label{fig:sensitivity}
\end{figure*}

\subsection{Sensitivity of \tuner~to $N$}
\label{app:sen}

We conduct a sensitivity analysis on the fixed number of evaluations $N$ for the generator. To that end, we set $N\in\{5,10,15\}$ From Figure~\ref{fig:sensitivity}, clearly, the empirical results are well-aligned with our theory: a too large $N$ would reduce the contribution from optimizing/exploring the retriever, while a too small $N$ can cause insufficient exploitation in optimizing the generator. Among others, $N=10$, which is our default, serves as a sweet point that leads to the best results, which we have set as default in this work.

\subsection{Holdout Test Queries Validation}
\label{app:holdout}

We evaluate the effectiveness of all methods on the holdout test queries from \textit{HotpotQA} and \textit{BioASQ}, using two backbone LLMs, \texttt{Llama-3.1-8B} and \texttt{Qwen-3-8B}. The \textit{Agriculture} and \textit{Biography} benchmarks are excluded due to the small sample size of their holdout sets, which is insufficient for reliable evaluation. To maintain consistency with the number of queries used during hyperparameter optimization, we construct the test sets by selecting an additional set of queries of matching size. For each method, we fix the best configuration obtained during optimization and test if under the holdout queries; we report the mean and standard deviation of F1 across 10 repeated runs. As shown in Tables~\ref{tab:rq1_main_two_models_two_datasets} and~\ref{tab:rq2_main_two_models_two_datasets}, the \tuner~variants consistently outperform their corresponding single-stage optimizers across both datasets and backbone models. Notably, the improvement is most pronounced on \textit{HotpotQA}, while gains on \textit{BioASQ} are smaller but consistent, demonstrating the robustness of the dual-stage optimization approach.

\section{Discussion}

\subsection{Precision vs. Richness in Retrieval Stage}

Indeed, when optimizing in the retrieval stage, there can be a trade-off between precision and richness, but their relationship is not strictly conflicting: there exist ``sweet point'' hyperparameter configurations that have both better precision and richness than the others, beyond which point biases to either might cause a significant drop in the generation quality. The dual-sequential formalization in \texttt{CDS4RAG} is designed to better find those ``sweet points'', and the cyclic iteration/refinement is purposely designed to implicitly mitigate the unwanted compromise in the trade-off: if in one cycle we bias too much towards precision (over-compromising richness), the generation quality under those related hyperparameter configurations would have worse ranks, hence making them less likely to influence the subsequent optimization.

\subsection{Ground Truth and Metrics}

\texttt{CDS4RAG} needs ground truth (for precision and F1-score), but we would like to kindly stress that this has been widely used in RAG hyperparameter optimization, given the vast types of benchmarks available. However, \texttt{CDS4RAG} is actually compatible with any metrics, which can be flexibly replaced. Practically, the suggested self-supervised or LLM-as-a-judge can be directly used in \texttt{CDS4RAG} when the ground truth is unavailable. 

Indeed, the token-level F1-score does not cover every aspect of RAG quality. We use it here because it is a widely-applied standard of the QA-style benchmarks for RAG, enabling fair comparison with prior work, and provides a consistent scalar final optimization target. Again, other metrics such as faithfulness and answer relevance can be seamlessly applied (note that the context precision has been used as the objective for hyperparameter optimization at the retrieval phase in \texttt{CDS4RAG}).

\section{Reproducibility}
All experiments use fixed random seeds for reproducibility. To support open science, the data, scripts, and full experimental results can be found in our anonymous repository at: \href{https://github.com/ideas-labo/cds4rag/}{https://github.com/ideas-labo/cds4rag/}.

\begin{table}[t!]
\centering
\small
\begin{adjustbox}{width=\linewidth}
\begin{tabular}{lcccc}
\toprule
\multirow{2}{*}{\textbf{Algorithm}} & \multicolumn{2}{c}{\textbf{Llama-3.1-8B}} & \multicolumn{2}{c}{\textbf{Qwen-3-8B}} \\
\cmidrule(lr){2-3} \cmidrule(lr){4-5}
& \textbf{HotpotQA} & \textbf{BioASQ} & \textbf{HotpotQA} & \textbf{BioASQ} \\
\midrule
\texttt{HEBO} & 0.500$_{\pm 0.075}$ & \textbf{0.265}$_{\pm \mathbf{0.017}}$ & 0.485$_{\pm 0.110}$ & 0.248$_{\pm 0.016}$ \\
\texttt{CDS4RAG(HEBO)} & \textbf{0.562}$_{\pm \mathbf{0.028}}$ & 0.261$_{\pm 0.013}$ & \textbf{0.594}$_{\pm \mathbf{0.036}}$ & \textbf{0.255}$_{\pm \mathbf{0.005}}$ \\
\midrule
\texttt{BO} & 0.462$_{\pm 0.125}$ & \textbf{0.255}$_{\pm \mathbf{0.020}}$ & 0.359$_{\pm 0.158}$ & 0.254$_{\pm 0.006}$ \\
\texttt{CDS4RAG(BO)} & \textbf{0.514}$_{\pm \mathbf{0.089}}$ & 0.240$_{\pm 0.014}$ & \textbf{0.602}$_{\pm \mathbf{0.032}}$ & \textbf{0.260}$_{\pm \mathbf{0.008}}$ \\
\midrule
\texttt{TPE} & 0.482$_{\pm 0.041}$ & 0.261$_{\pm 0.016}$ & \textbf{0.553}$_{\pm \mathbf{0.117}}$ & 0.245$_{\pm 0.020}$ \\
\texttt{CDS4RAG(TPE)} & \textbf{0.544}$_{\pm \mathbf{0.025}}$ & \textbf{0.273}$_{\pm \mathbf{0.011}}$ & 0.537$_{\pm 0.042}$ & \textbf{0.254}$_{\pm \mathbf{0.010}}$ \\
\bottomrule
\end{tabular}
\end{adjustbox}
\caption{RQ1 comparison between \texttt{CDS4RAG} and corresponding single-stage algorithms. Each cell represents mean$_{\pm\text{std}}$ of F1-score.The bold indicates the better algorithm of each case.}
\label{tab:rq1_main_two_models_two_datasets}
\end{table}

\begin{table}[t!]
\centering
\small
\begin{adjustbox}{width=\linewidth}
\begin{tabular}{lcccc}
\toprule
\multirow{2}{*}{\textbf{Algorithm}} & \multicolumn{2}{c}{\textbf{Llama-3.1-8B}} & \multicolumn{2}{c}{\textbf{Qwen-3-8B}} \\
\cmidrule(lr){2-3} \cmidrule(lr){4-5}
& \textbf{HotpotQA} & \textbf{BioASQ} & \textbf{HotpotQA} & \textbf{BioASQ} \\
\midrule
\texttt{CDS4RAG(HEBO)} & \textbf{0.562$_{\pm \mathbf{0.028}}$} & \textbf{0.261$_{\pm \mathbf{0.013}}$} & \textbf{0.594$_{\pm \mathbf{0.036}}$} & \textbf{0.255$_{\pm \mathbf{0.005}}$} \\
\texttt{RayTune} & 0.482$_{\pm 0.041}$ & 0.261$_{\pm 0.016}$ & 0.553$_{\pm 0.117}$ & 0.245$_{\pm 0.020}$ \\
\texttt{Random} & 0.502$_{\pm 0.049}$ & 0.254$_{\pm 0.018}$ & 0.476$_{\pm 0.115}$ & 0.250$_{\pm 0.012}$ \\
\texttt{Greedy} & 0.498$_{\pm 0.110}$ & 0.237$_{\pm 0.016}$ & 0.448$_{\pm 0.145}$ & 0.246$_{\pm 0.019}$ \\
\texttt{AutoRAG-HP} & 0.426$_{\pm 0.069}$ & 0.244$_{\pm 0.013}$ & 0.410$_{\pm 0.164}$ & 0.242$_{\pm 0.017}$ \\
\bottomrule
\end{tabular}
\end{adjustbox}
\caption{RQ2 comparison with other state-of-the-art algorithm for RAG hyperparameter optimization. Each cell represents mean$_{\pm\text{std}}$ of F1-score.The bold indicates the better algorithm of each case.}
\label{tab:rq2_main_two_models_two_datasets}
\end{table}